\documentclass[conference]{IEEEtran}

\usepackage{url}
\usepackage{CJKutf8}
\usepackage{tabularx}
\usepackage{makecell}
\usepackage{amsmath}
\usepackage{amsfonts}
\usepackage{xcolor}
\usepackage{xspace}
\usepackage{array}
\usepackage{hyperref}
\usepackage{booktabs}

\usepackage{algorithmicx}
\usepackage{indentfirst}
\usepackage{graphicx}
\usepackage{diagbox}
\usepackage{arydshln}
\usepackage{multirow}
\usepackage[center]{caption}
\usepackage{subcaption}
\usepackage{amsthm}
\usepackage[abbreviations]{foreign}

\usepackage{easyReview}

\usepackage{float}
\usepackage{afterpage}
\usepackage[shortlabels]{enumitem}
\usepackage[justification=justified,format=plain]{caption}
\usepackage{pifont}
\usepackage{cite}
\newcommand{\cmark}{\ding{51}}
\newcommand{\xmark}{\ding{55}}

\newcommand{\gray}[1]{\textcolor{gray}{#1}}

\newcommand*{\SuperScriptSameStyle}[1]{%
  \ensuremath{%
    \mathchoice
      {{}^{\displaystyle #1}}%
      {{}^{\textstyle #1}}%
      {{}^{\scriptstyle #1}}%
      {{}^{\scriptscriptstyle #1}}%
  }%
}
\newcommand*{\oneS}{\SuperScriptSameStyle{*}}
\newcommand*{\twoS}{\SuperScriptSameStyle{**}}
\newcommand*{\threeS}{\SuperScriptSameStyle{*{*}*}}
\DeclareMathOperator*{\var}{Var}
\DeclareMathOperator*{\cov}{Cov}

\newcommand{\reg}[1]{$\mathbb{#1}$}

\let\OLDitemize\itemize
\renewcommand\itemize{\OLDitemize\setlength{\itemsep}{0pt}}

\setlist[itemize]{noitemsep, topsep=0pt,wide=\parindent}
\setlist[enumerate]{wide=\parindent, noitemsep, topsep=0pt}
\renewcommand{\figurename}{Figure}
\renewcommand{\tablename}{Table}
\newcommand{\ignore}[1]{}

\newcolumntype{s}{>{\centering\hsize=.5\hsize}X}

\theoremstyle{plain}
\newtheorem{observation}{Observation}

\theoremstyle{definition}
\newtheorem{definition}{Definition}
\theoremstyle{remark}

\theoremstyle{Proposition}
\newtheorem{proposition}{Proposition}

\begin{document}
\title{Transfer Attacks Revisited: A Large-Scale Empirical Study in Real Computer Vision Settings}

\author{
  \IEEEauthorblockN{Yuhao Mao\IEEEauthorrefmark{1}\IEEEauthorrefmark{2}, Chong Fu\IEEEauthorrefmark{1}, Saizhuo Wang\IEEEauthorrefmark{1}, Shouling Ji\IEEEauthorrefmark{1}(\ding{41}), Xuhong Zhang\IEEEauthorrefmark{1}\IEEEauthorrefmark{3}(\ding{41}), \\ Zhenguang Liu\IEEEauthorrefmark{7},  Jun Zhou\IEEEauthorrefmark{4},  Alex X. Liu\IEEEauthorrefmark{4},  Raheem Beyah\IEEEauthorrefmark{5}, Ting Wang\IEEEauthorrefmark{6}}
\IEEEauthorblockA{\IEEEauthorrefmark{1}Zhejiang University,  \IEEEauthorrefmark{2}ETH Zürich, \IEEEauthorrefmark{3}Zhejiang University NGICS Platform, \IEEEauthorrefmark{7}Zhejiang Gongshang University, \IEEEauthorrefmark{4}Ant Group, \\ \IEEEauthorrefmark{5}Georgia Institute of Technology,  \IEEEauthorrefmark{6}Pennsylvania State University} 

\{yuhaomao, fuchong, szwang, sji, zhangxuhong\}@zju.edu.cn, liuzhenguang2008@gmail.com, zhoujun@antfin.com, \\ alexliu@antgroup.com, rbeyah@ece.gatech.edu, inbox.ting@gmail.com
}

\maketitle

\begin{abstract}
    One intriguing property of adversarial attacks is their ``transferability'' -- an adversarial example crafted with respect to one deep neural network (DNN) model is often found effective against other DNNs as well. Intensive research has been conducted on this phenomenon under simplistic controlled conditions. 
    Yet, thus far there is still a lack of comprehensive understanding about transferability-based attacks (``transfer attacks'') in real-world environments.

    To bridge this critical gap, we conduct the first large-scale systematic empirical study of transfer attacks against major cloud-based MLaaS platforms, taking the components of a real transfer attack into account. The study leads to a number of interesting findings which are inconsistent to the existing ones, including:
    ({\em i}) Simple surrogates do not necessarily improve real transfer attacks.
    ({\em ii}) No dominant surrogate architecture is found in real transfer attacks.
    ({\em iii}) It is the gap between posterior (output of the softmax layer) rather than the gap between logit (so-called $\kappa$ value) that increases transferability.
    Moreover, by comparing with prior works, we demonstrate that transfer attacks possess many previously unknown properties in real-world environments, such as 
    ({\em i}) Model similarity is not a well-defined concept.
    ({\em ii}) $L_2$ norm of perturbation can generate high transferability without usage of gradient and is a more powerful source than $L_\infty$ norm.
    We believe this work sheds light on the vulnerabilities of popular MLaaS platforms and points to a few promising research directions. \footnote{Code \& Results: https://github.com/AlgebraLoveme/Transfer-Attacks-Revisited-A-Large-Scale-Empirical-Study-in-Real-Computer-Vision-Settings}
\end{abstract}

\section{Introduction}

Deep neural networks (DNNs) achieve tremendous success in a variety of application domains \cite{he_deep_2016, vaswani_attention_2017}, but they are inherently vulnerable to adversarial examples (AEs) which are malicious samples crafted to deceive target DNNs \cite{szegedy2013intriguing, DBLP:journals/corr/abs-2110-06018, DBLP:journals/corr/abs-2202-10673, DBLP:journals/corr/abs-2012-09302}. This vulnerability significantly hinders their use in security-sensitive domains.

Among the many properties of AEs, the transferability, that an AE crafted with respect to one DNN also works against other DNNs, is particularly intriguing. Leveraging this property, the adversary could forge AEs using a surrogate DNN to attack the target DNN without knowledge of the target. This is highly dangerous because an increasing number of cloud-based MLaaS platforms are deploying DNNs with public API.
Furthermore, given that years' advances in technology did not eliminate this vulnerability, it is highly likely that
transferability stems from the intrinsic properties of DNNs. Understanding the phenomenon improves the interpretability of DNNs in its own right as well.

\begin{figure}
    \centering
    \begin{subfigure}{0.4\linewidth}
        \centering
        \includegraphics[width=0.5\textwidth]{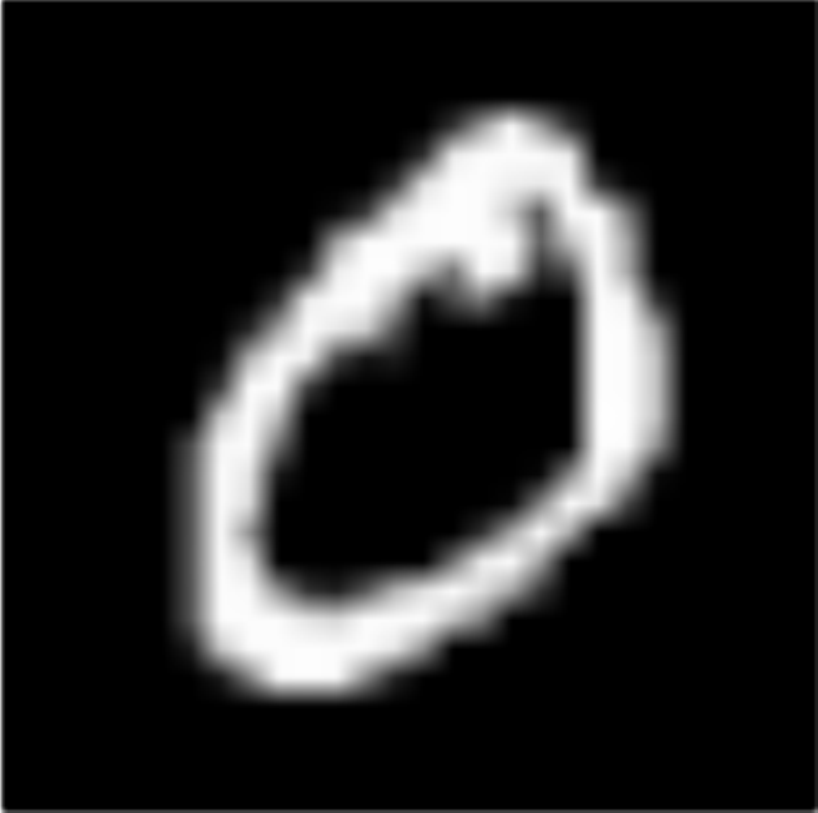}
        \caption{\small 97.2\% Text; 96.5\% Number; 96.5\% Symbol.}
        \label{subfig:academic_image}
    \end{subfigure}
    \begin{subfigure}{0.4\linewidth}
        \centering
        \includegraphics[width=0.5\textwidth]{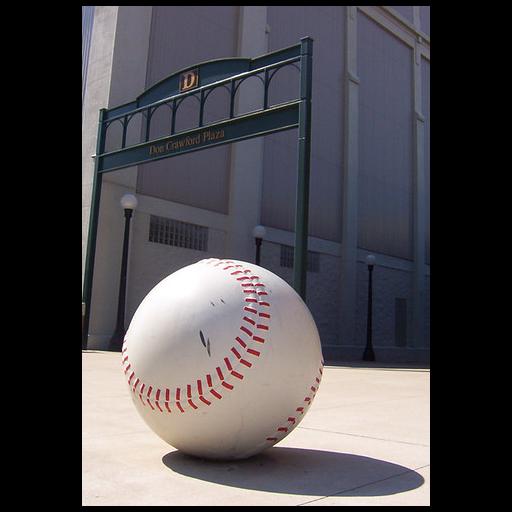}
        \caption{\small 91.7\% Sports; 86.7\% Sphere; 78.9\% Baseball.}
        \label{subfig:motivation_b}
    \end{subfigure}
    \caption{AWS Rekognition's predictions to two sample images. The left one does not contain the desired prediction (digit zero) and the right one's ground truth label (baseball) does not get the highest score. They are not mistakes but metrics like top-k accuracy would consider them as not good.}
    \label{fig:motivation}

\end{figure}

Therefore, understanding the deciding factors of transferability and their working mechanisms has attracted intensive researches \cite{szegedy2013intriguing,
    papernot_transferability_2016, goodfellow2014explaining, liu2016delving,
    demontis2018adversarial, DBLP:conf/kdd/PangZJLW20, DBLP:conf/ccs/ShenJ0LCSFYW21}. However, to the best of our knowledge, all systematic empirical studies are conducted under controlled ``lab'' environments that are too ideal to make the derived conclusions reliable in the real environment. For instance, many
studies \cite{papernot_practical_2017, papernot_transferability_2016} give the adversary access to the training data of the target
model, and some others \cite{su2018robustness} discuss surrogates with similar complexity to the target model.
Although there are studies that attack deployed DNNs known as cloud models \cite{liu2016delving}, their conclusions are neither systematic nor informative enough, suffering from the lack of adequate observations to do statistical tests and suitable metrics to account for the difference between lab and real settings.
The difference between lab and real environment includes:
\begin{itemize}
    \item \textbf{Complexity and architecture of the target.}
          In the real environment, neither the target's complexity nor its architecture is known to the attacker. The cloud model could be arguably far more complicated than a surrogate and of an uncommon architecture.
          Under this circumstance, the conclusions derived from academic target models may not hold, calling for thorough examinations.
    \item \textbf{Training of the target.}
          In the real environment, the adversary knows nothing about the training
          details of the target, including the optimization hyperparameters and algorithms. Additionally, the training datasets are far more complicated and noisy in the real environment\footnote{Amazon declares they have access to billions of images daily and continue to learn from new data \cite{aws_doc}.}, increasing the difficulty in simulating a similar training environment at
          local. 
    \item \textbf{Structure of the input.}
          The cloud models are designed for high resolution images. Therefore, the cloud models
          may not produce meaningful outputs for inputs from academic datasets that are extremely low resolution like MNIST \cite{mnist} and CIFAR10 \cite{cifar}, as shown in Figure \ref{subfig:academic_image}.
          Conclusions made on these toy datasets may not hold under this condition.

    \item \textbf{Structure of the output.}
          Cloud models usually return multiple predictions and their corresponding confidences, which are different from the logits returned by classifiers in lab settings (\emph{multiple returns problem}). In addition, cloud models usually have a significantly larger set of labels\footnote{Google shares an open dataset consisting of about 60M labels \cite{googleapis}.} that are hard to estimate and categorize locally (\emph{class inconsistency problem}). These differences, as depicted in Figure~\ref{fig:motivation}, add to the difficulty in measuring the effectiveness of transfer attacks in a real-world scenario.
\end{itemize}

To address these limitations, we conduct a large-scale systematic empirical study on
transfer attacks in the real settings. To comprehensively explore the possible factors
that may affect the transferability, we combine different settings of the attack
components at local environment where  AEs are generated. To thoroughly evaluate
the effectiveness of transfer attacks and the robustness of real-world models, we
select four leading cloud-based Machine-Learning-as-a-Service (MLaaS)
platforms, namely Aliyun (Alibaba Cloud), Baidu Cloud, AWS Rekognition and Google Cloud Vision as the targets.

Our contributions include:

\begin{itemize}
    \item We identify the main gaps of evaluating transfer attack in the real world, namely \emph{multiple returns problem} and \emph{label inconsistency problem}, and extend existing metrics to address these difficulties. To the best of our knowledge, these are the first metrics that can be reasonably applied to analyze transfer attack on MLaaS systems.
    \item We conduct a systematic evaluation on the transferability of adversarial
          attacks against four leading commercial MLaaS platforms on two computer
          vision tasks: object classification and gender classification, using datasets
          consisting of real-world pictures in order to get meaningful insights (ImageNet \cite{deng2009imagenet} and Adience \cite{adience}, respectively). Based on the results of our evaluation, we measure the robustness of the discussed platforms and point out that while on average transfer attack performs poorly, they can be systematically designed to achieve stable success rates, \eg, use FGSM for the object classification.
           More detailed guidelines for the industry is provided in Appendix \ref{sec:guideline}. We hope this will help the
          industry fortify their models and improve their services.
    \item We explore the possible factors that may affect the transferability using
          180 different settings as the combination of components in a transfer attack. 36,000 AEs in total generated from 200 seed images for each task are evaluated, making the conclusions statistically adequate. We find that some former conclusions can be
          generalized to the real settings while some others are inapplicable. A
          comparison between the former conclusions and their real-world counterparts is
          shown in Table \ref{tb:comparison_of_conclusions}. Additionally, we demonstrate that transfer attacks possess many previously unknown properties in real-world environments. We believe that our
          findings will provide new insights for future
          research.
\end{itemize}

\begin{table*}[htb]
    \centering
    \caption{Conclusions comparison between lab settings and real settings. }
    \begin{tabularx}{\textwidth}{p{6.6cm}p{8.6cm}p{0.3cm}}
        \toprule
        \multicolumn{1}{c}{\textbf{Former conclusion in labs}}                                                      & \multicolumn{1}{c}{\textbf{Refined conclusion in reality}}                                                                                                                                                                                                                        & \multicolumn{1}{c}{\textbf{Consistency}} \\ \midrule

        Untargeted Attacks are easy to transfer and targeted attacks almost never transfer. \cite{liu2016delving}   & All kind of transfer attacks discussed has a significantly positive success rate.
                                                                                                                    & \multicolumn{1}{c}{Partial}
        \\

        \emph{FGSM} $\ge$ \emph{PGD} $\ge$ \emph{CW}
        in the sense of transferability. \cite{su2018robustness}                                                    & (1) Targeted algorithms are worse than untargeted algorithms in both targeted and untargeted transfer attack. (2) Single-step algorithms are better than iterative algorithms.
                                                                                                                    & \multicolumn{1}{c}{\cmark}
        \\

        Surrogates with less complexity, defined by smaller variability of the loss landscape, are better choices for transfer attack. Experimentally, a surrogate with simpler structure and stronger regularization is better. \cite{demontis2018adversarial}                    &
        The effect of surrogate complexity, defined by the number of layers and parameters of the neural network, is non-monotonic. A surrogate with a good depth can outperform simpler and deeper surrogates.
                                                                                                                    & \multicolumn{1}{c}{\xmark}                                                                                                                                                                                                                                                                                                   \\

        AEs crafted from VGG surrogates transfer well to all targets, while others almost never transfer to targets in different families.  \cite{su2018robustness}
                                                                                                                    & No dominant architecture is found to craft AEs that transfer better to cloud models.
                                                                                                                    & \multicolumn{1}{c}{\xmark}
        \\

        Relaxing $L_\infty$ norm perturbation constraint largely increases transferability. \cite{su2018robustness} & $L_2$ norm of perturbation has far stronger correlation with transferability. Increasing $L_2$ norm while keeping $L_\infty$ fixed makes good transferability but large $L_\infty$ norm with small $L_2$ norm can give poor success rate. 
                                                                                                                    & \multicolumn{1}{c}{Partial}                                                                                                        \\

        A larger logit gap between the adversarial class and the second likely class, called $\kappa$ value, increases untargeted attack transferability.  \cite{carlini2016evaluating} \cite{su2018robustness}
                                                                                                                    & The gap between the posterior of logits is a better representative for measuring transferability. A large $\kappa$ cannot  increase transferability in many cases.                                                                                                                                                         &
        \multicolumn{1}{c}{\xmark}
        \\
        \bottomrule
    \end{tabularx}
    \label{tb:comparison_of_conclusions}
\end{table*}

\section{Background}
\label{sec:backgrounds}
In this section, we introduce the preliminaries of transfer attacks, including
the components that may affect the transferability. 

\subsection{Transfer Attack}
Transfer attack involves \textit{adversarial attack} and
\textit{transferability}. Specifically, a classifier with parameters
$\theta$, namely $f_\theta$, accepts an input $x \in R^n$ and makes a prediction
$f_\theta(x)$. The
adversary wants to generate a maliciously perturbed
input $\hat{x} = x + \delta$ so that $f_\theta(\hat{x})\ne f_\theta(x)$.
Typically, $\delta$ is required to be human imperceptible, \ie, the size of $\delta$ is small under $L_p$ norm.
Attack algorithms of this form are called \textit{adversarial attacks} \cite{szegedy2013intriguing, goodfellow2014explaining, carlini2016evaluating, papernot_limitations_2016, moosavidezfooli2015deepfool, aleks2017deep, tramr2017ensemble, DBLP:journals/tdsc/LiJHJRLW21}.

Formally, if an AE
$\hat{x}$ crafted on $f_\theta$ deceives another model
$\phi_{\theta ^\prime}$ with unknown parameters and architecture, \ie, $
\phi_{\theta ^\prime}(\hat{x}) \ne \phi_{\theta ^\prime}(x)$, then we say this AE
\textit{transfers} from model $f_{\theta}$ to $\phi_{\theta ^\prime}$. Utilizing this
property, an adversary may generate AEs based on $f_\theta$ and transfer them to
the target black-box model $\phi_{\theta ^\prime}$. Typically, the label sets of the source model and the target model are the same. This process is called
\textit{transfer attack} \cite{papernot_transferability_2016, papernot_practical_2017, liu2016delving}.
Although query-based attacks \cite{chen_zoo_2017, ilyas_black-box_2018} that request partial information about the targets are more powerful, transfer attacks do not require any information about the target and thus are more stealthy and economic.


According to the process of transfer attack, there are three major components affecting the transferability of AEs: \textit{surrogate model}, \textit{surrogate dataset} and \textit{adversarial algorithm}. 
We elaborate on these components next.

\subsection{Surrogate Model}
\label{sec:surrogate-model}
Surrogate model is the model on which the adversary performs white-box attacks. In general, it is expected to resemble the target model to increase the transferability of the generated AEs. The main factors affecting the similarity include:
\begin{enumerate}[(a)]
    \item \textit{Pretraining}. 
    Target models, such as the ones on MLaaS platforms, are trained on
    a sufficiently large dataset to achieve a superior performance. Since 
    attackers may not have enough computing resources to train the
    surrogate model on a large dataset, they may
    fine-tune the surrogate model based on a public model that is
    pretrained on a large dataset, as pretraining generally improves a model's accuracy  
    \cite{devlin-etal-2019-bert, NIPS2019_xlnet}.

    \item \textit{Model Architecture.} 
    Various DNN architectures have been designed to address different problems
    and improve performance \cite{simonyan_very_2015, he_deep_2016, googlenet}.
    The fitting functions that models learned under different architectures could be different even with the same training data, which might influence the transferability of generated AEs.
    
    \item \textit{Model Complexity.} 
    For one specific architecture, models with different complexities usually have different performance. This phenomenon indicates that they have different capabilities in capturing the distribution of the training data, thus affecting the transferability of AE when used as surrogate models. In this paper, since we are in the context of neural networks, we define the model complexity by the number of layers and parameters. On the contrary, Demontis \etal \cite{demontis2018adversarial} define the model complexity by the variability of the loss landscape and the input gradient size. We experimentally confirm their conclusions on the input gradient size on local targets because the gradients of the MLaaS models are non-transparent. For more details, please refer to Appendix \ref{appendix:input_grad}. The variability of the loss landscape is computationally expensive for the large surrogates we use, thus we do not use this as the complexity metric.

\end{enumerate}

\subsection{Surrogate Dataset}
\label{sec:surrogate-dataset}
Surrogate dataset is the dataset used to train the surrogate model. In general,
the distribution of the surrogate data is expected to approximate the distribution
of the data used to train the target. In this way, the similarity between the surrogate and the target model is expected to increase so that the AEs crafted on the surrogate are easier to transfer. However, the surrogate datasets that an attacker possesses are usually much smaller than those owned by the MLaaS systems. To make the surrogate dataset capture the distribution of the target model's training data as well as possible, there are two common data enrichment methods:
\begin{enumerate}[(a)]
    \item \textit{Data Augmentation.} 
    Data augmentation is to apply transformations on an image while maintaining the primary patterns critical to the classification. Applying augmentation enforces a model to learn patterns under extended situations and focus on the dominant patterns.
    
    \item \textit{Adversarial Training.} 
    Adversarial training is to enrich the training data with their adversarial
    counterparts so that the surrogate
    model can focus on robust features which are not changeable by small
    perturbations \cite{EAT2018, kurakin2016scale}.
\end{enumerate}

These methods allow the surrogate to learn better features from the limited surrogate dataset.

\subsection{Adversarial Algorithm}
\label{sec:wba}
Adversarial algorithm is the white-box attack algorithm (WBA) used to generate AEs on the surrogate model. 
Among the WBAs, there are several common properties separating them into different categories:
\begin{enumerate}[(a)]
    \item Depending on the goal, WBAs can be categorized into \textit{targeted} and \textit{untargeted}. Targeted algorithms aim to let the model predict the target class, \ie, $f_\theta(\hat{x}) = c_t$, where $c_t$ denotes the target class. Untargeted algorithms merely want the model to misclassify, \ie, $f_\theta(\hat{x}) \neq c_o$, where $c_o$ denotes the original predicted class.

    \item Depending on the optimization process, WBAs can be categorized into \textit{single-step attacks} which add perturbations to the original input only once, and \textit{iterative attacks} which perturb the original input repeatedly until a certain condition is satisfied.

\end{enumerate}

\section{Threat Model}
\label{sec:threat_model}
\begin{figure}
    \centering
    \includegraphics[width=.8\linewidth]{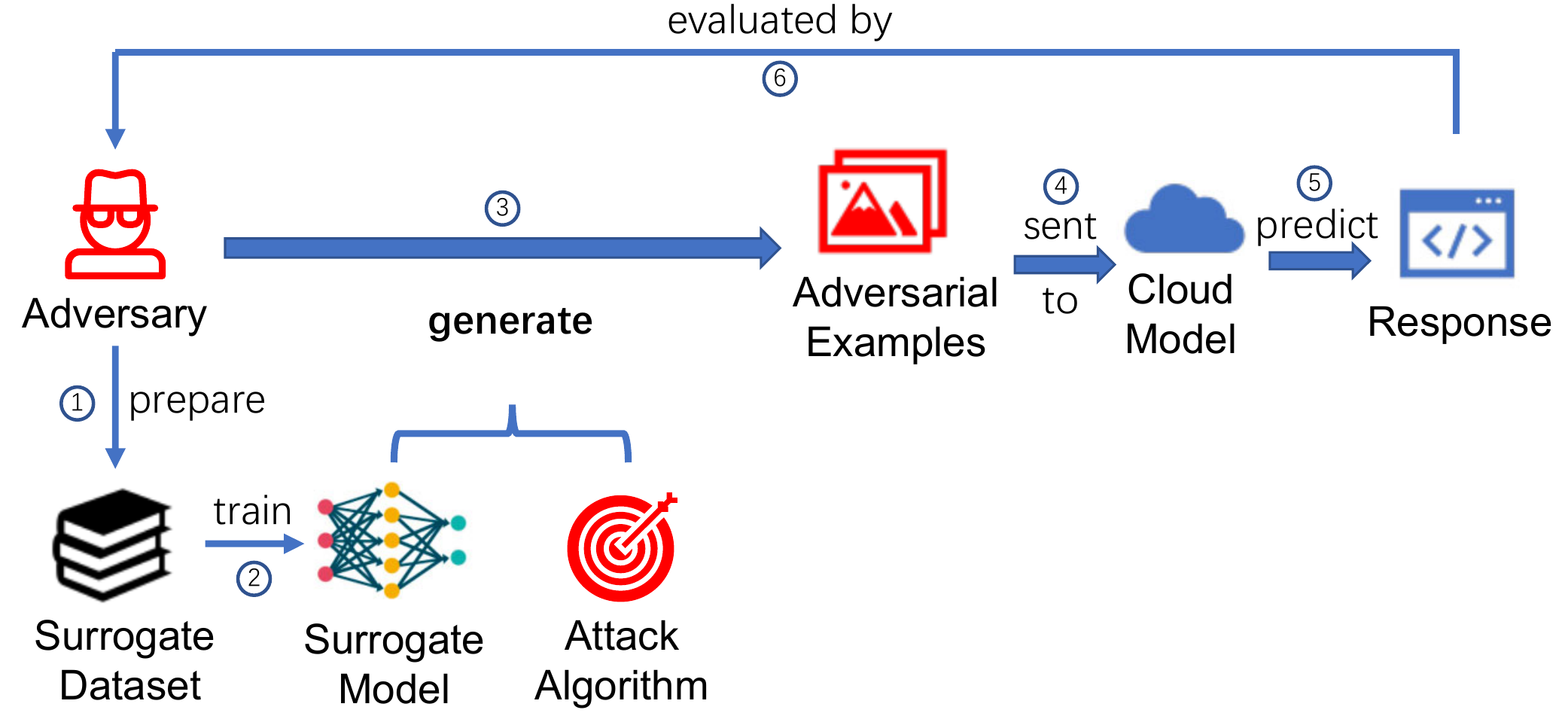}
    \caption{Pipeline of transfer attack against MLaaS platforms.}
    \label{fig:pipeline}
\end{figure}

Figure~\ref{fig:pipeline} provides an overview of transfer attacks on MLaaS
platforms. The model structure and training dataset of the target platform model is
obscure to an attacker. Additionally, an attacker can only access the target
platform model by sending images to the MLaaS
platform and getting their predictions via
the platform's API. Therefore, an attacker can only manipulate the input of the target platform model to perform an attack. 
For ease of understanding, we formally present the following
attack-related definitions: 
(1) \textit{targeted AE} is defined as the AE
generated locally via targeted algorithms;
(2) \textit{untargeted AE} is
defined as the AE generated locally via untargeted algorithms;
(3) \textit{targeted transfer attack} is defined as the targeted attack against the
MLaaS platforms using locally generated AEs;
(4) \textit{untargeted transfer attack} is defined as
the untargeted attack against the MLaaS platforms using locally generated AEs;

Following Chen \etal \cite{chen2020universal}, we consider the case that attackers maintain a pool of highly transferable AEs as well. Under such condition, we are actually viewing transferability as a property of AEs and every AE can be used to launch any kind of transfer attack. It immediately leads to the following two key differences:
(1) Both targeted and untargeted AEs can
    be used to perform targeted and untargeted transfer attacks. For targeted transfer attacks, we take the predicted label of an AE by the surrogate as the target label.
(2) The correlation between transferability and sample-level properties such as adversarial confidence and perturbation size can be easily examined.

\section{Evaluation Settings and Metrics}
\subsection{Evaluation Settings}
\label{sec:specific-settings}
\subsubsection{Surrogate Model Settings}
\label{sec:surrogate-model-set}
For each surrogate model, we evaluate multiple settings for each of its
components discussed in Section \ref{sec:surrogate-model}. We use the
implementations of models in PyTorch \cite{pytorch} and
TorchVision \cite{torchvision} libraries to conduct our experiments. For different model complexities, we apply
three ResNet models with different depths: ResNet-18, ResNet-34 and ResNet-50. These depths are chosen because they are officially implemented.
We apply Inception V3 \cite{szegedy_rethinking_2016},
VGG-16 \cite{simonyan_very_2015} and ResNet-18 \cite{he_deep_2016} to study the impact of architecture. 
For pretraining, we compare the raw ResNet models without pretraining and the ResNet
models pretrained on the whole ImageNet dataset. For the object classification task, we only finetune the last fully connected layer and fix other layers. For the gender classification task, we use the pretrained parameters as an initialization and finetune the whole model. Empirical result shows all pretrained surrogates have a significant improvement on accuracy. Table
\ref{tb:setting_component} summarizes each component.

\begin{table}
  \centering
  \caption{Settings for each component of the surrogate model.}
  \begin{tabular}{cc}
    \hline
    \textbf{Component} & \textbf{Settings}         \\
    \hline
    Architecture       & Inception, VGG, ResNet    \\
    Complexity         & (ResNet Depth) 18, 34, 50 \\
    Pretraining         & True, False               \\
    \hline
  \end{tabular}
  \label{tb:setting_component}
\end{table}

\subsubsection{Surrogate Dataset Settings}
\label{sec:surrogate-data-set}
For the image classification task, we use a subset of ImageNet \cite{deng2009imagenet} consisting of 10 classes.
Each class contains approximately 500 images cropped to 512$\times$512$\times$3. It is not an important reduction in the number of samples because these classes have roughly 500-600 images per class in the original ImageNet data. We do this because the number of targets' classes is less meaningful in attacking, while the structure of the data is critical, as shown in Figure \ref{fig:motivation}. Training the surrogate with a subset is to lower the cost and have a better simulation of real attacks with limited resources.
Similarly, for gender classification, we use a subset of the Adience dataset \cite{adience}.
We randomly pick 10,000 images from the Adience dataset with half male and half female and crop them to 384$\times$384$\times$3.
For each of the two surrogate datasets, we randomly split it into training set, validation set and test set with the split ratio  8:1:1.

We apply the two data enlargement techniques discussed in Section \ref{sec:surrogate-dataset}.
Specifically, for data augmentation, we use color jitter, random affine, random horizontal flip, random perspective, random rotation and random vertical flip. They are imposed on an image independently with a probability of 0.5. The detailed parameter settings can be found in Appendix \ref{appendix:params}.
For adversarial training, we employ the naive adversarial training (NAT) algorithm \cite{kurakin2016scale}.
Thus, we have three variants for each dataset: \textit{raw} (the original dataset), \textit{augmented} and \textit{adversarial}.

\subsubsection{Adversarial Algorithm Settings}
\label{def:attack_group}

In total, we employ nine representative adversarial attacks.
Among them there are two targeted algorithms,  BL-BFGS (simplified as BLB) \cite{szegedy2013intriguing} and CW2 \cite{carlini2016evaluating}.
BLB and CW2 generate AEs by solving the corresponding optimization problem
using iterative methods. Adversarial perturbations generated by them are optimized in size and thus by nature small.
Apart from these targeted algorithms, we employ six untargeted adversarial algorithms as well. DeepFool \cite{moosavidezfooli2015deepfool} is the only untargeted algorithm that optimize the perturbation size, realized by approximating the decision boundary. A generalized version of DeepFool is the UAP algorithm \cite{moosavidezfooli2016universal}, adding image-level adversarial perturbations given by DeepFool together to obtain a universal adversarial perturbation. Algorithms remained are built upon another idea that AEs should be generated with $L_p$ norm of perturbation less than a predetermined norm budget. The
root algorithm is FGSM \cite{goodfellow2014explaining} which perturbs the
original image for once along the opposite direction of the gradient,
\ie, $\hat{x} = x - \epsilon \cdot sign(\nabla loss(f_\theta(x); c)$, where
$loss(f_\theta(x); c)$ is the loss function between the model prediction and the
ground truth, $sign(\cdot)$ denotes the sign function and $\epsilon$ denotes the perturbation budget. FGSM has its improved version
RFGSM \cite{tramr2017ensemble} which adds a random perturbation to the image
before imposing gradient-based perturbation. This preprocessing is claimed to 
penetrate the defense of gradient masking. In addition, FGSM has an iterative
counterpart PGD \cite{aleks2017deep} which executes the FGSM step iteratively
for predetermined times. Another
variation of FGSM called Step-LLC and its iterative counterpart
LLC \cite{kurakin2016adversarial} minimize the loss w.r.t. the
least likely class rather than maximize the loss w.r.t. the correct class. The perturbing process then
becomes $\hat{x} = x + \epsilon \cdot sign(\nabla loss(f_\theta(x); c^*)$, where
$c^*$ denotes the class the model predicts with the least confidence. These attacks have $L_p$ norm of adversarial perturbation less than or equal to the budget. We use $L_\infty$ norm in the paper.

For algorithm implementations, we employ the codes from the open source framework DEEPSEC \cite{ling_deepsec_2019}.
We select ten independent classes from the ImageNet and randomly sampled 20 images for each class as the seed image.
For gender classification, we randomly select 200 original images as the seed image, half male and half female. The combinations of surrogate settings and attack algorithms form 180 different settings in total for our evaluation.
Under each setting, AEs are generated on the same set of seed images and sent to the target platform. Then metrics are computed based on the responses. 
\subsubsection{Cloud Experiment Settings}
We conduct our experiments on four leading commercial MLaaS platforms: Google Cloud Vision \cite{GCV}, AWS Rekognition \cite{AWS}, Aliyun (Alibaba Cloud) \cite{Aliyun} and Baidu Cloud \cite{Baidu}.
We send the AEs to these clouds using the official API provided by each platform and save the responses for evaluation.\footnote{We exclude Google for gender classification because it does not support this function.}

\subsection{Evaluation Metrics}
\label{sec:metric}

It is not trivial to determine whether a transfer attack is successful in real settings
for the multi-class classification tasks like ImageNet.
Basically, a MLaaS platform's response to a certain input image can be denoted as $P = \{(l, s)|l \in L, s \in [0, 1]\}$, where $l$ represents a label from the class set $L$ on the MLaaS platform  and $s$ denotes the confidence score for the prediction $l$.
Consequently, two challenging problems come up in matching a local class with the response from a MLaaS platform:
\begin{enumerate}[(a)]
  \item \textit{Label inconsistency problem}:
  Since $L$ is significantly different from the local class set $C$, $l$ cannot be directly associated with a local label $c$ because $l$ can either be a sub-class or super-class of $c$.
  For example, the local class ``weapon'' may correspond to classes in $L$ that are sub-classes of it, such as  ``gun'' and ``knife'', and the local class ``baseball'' may be predicted as a super-class, such as ``sport''.
  Additionally, $L$ differs across MLaaS platforms as well. 

  \item \textit{Multiple predictions problem}:
  While $P$ consists of multiple predictions, the ground truth is a single label. Hence, it is difficult to fairly judge whether the MLaaS platform is correct, since the expected predication might have a fairly high score but is not a top-$k$ label, as what we have seen in Figure \ref{subfig:motivation_b}.
  As a result, metrics like top-$k$ accuracy can not be reasonably applied because it only considers the prediction with the top $k$  scores.
\end{enumerate}

To address the label inconsistency problem, we construct a class mapping $M_c = \{l_c|l_c \in L\}$ for each local class $c$ and each MLaaS platform. The content of the constructed mapping can be found in the code repository. To construct $M_c$, we inspect all the responses of a MLaaS platform to the original image belonging to class $c$ and manually select the relevant classes from the responses. We then build an equivalence dictionary $T = \{(c, M_c)| c \in C\}$ for each MLaaS platform.
 We discuss the validity and potential bias introduced by the equivalence dictionary in Appendix \ref{appendix:dict_justify}.

\begin{figure}
  \centering
  \includegraphics[width=0.5\linewidth]{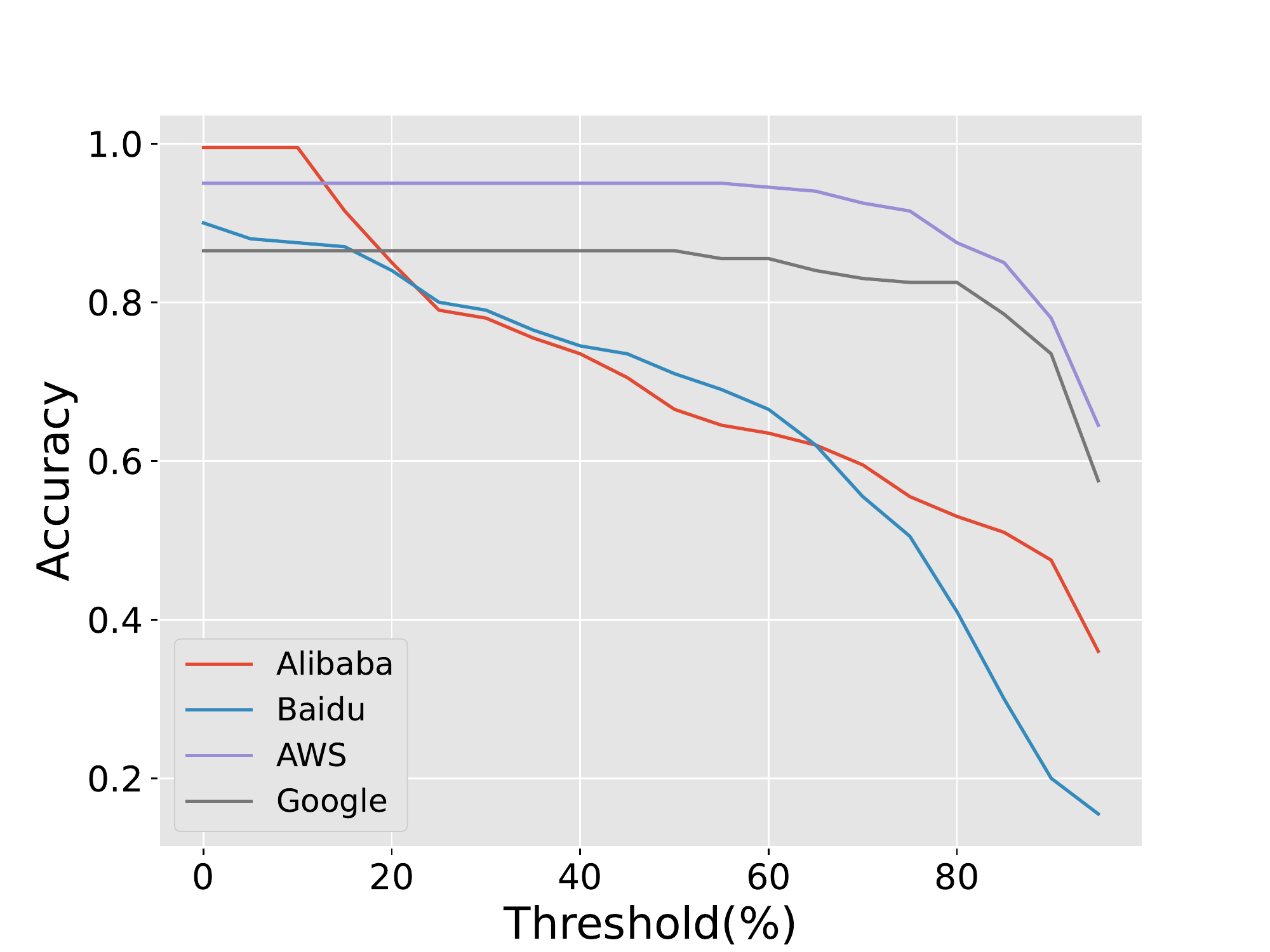}
  \caption{The accuracies of clouds with respect to different confidence thresholds.}
  \label{fig:threshold-accuracy}
\end{figure}

To address the multiple predictions problem, we select a confidence threshold $\sigma$ for each platform to filter out the predictions with low scores.
The value of the threshold $\sigma$ is critical to the fairness of our evaluation and should be meticulously chosen.
Our principle is that the prediction accuracy for clean images should not drop significantly and low confidence predictions are ruled out as many as possible.
To set an appropriate threshold for each cloud platform, we measure the prediction accuracies of the studied cloud platforms on the original ImageNet data under different thresholds.
As shown in Figure \ref{fig:threshold-accuracy}, the threshold value largely
affects the prediction accuracy and the trend varies across platforms. We find that Google and AWS mainly set high scores for their responded predictions, while Alibaba and Baidu attach low scores to many responded predictions. Therefore, we use  $\sigma=50\%$ for AWS and Google and $\sigma=10\%$ for Alibaba and Baidu. The difference in threshold mitigates the unfairness of setting a global confidence threshold to evaluate their robustness and leaves the analysis of other factors,  assumed to be independent to the target model, unharmed. More discussions about threshold cutting is provided in Appendix \ref{appendix:threshold}.

Based on the discussions above, the matching between a local class $c$ and a response $P$ from a MLaaS platform can be formalized as: $c$ match $P \Leftrightarrow \exists (l, s) \in P, s > \sigma \land l \in M_c$, where $M_c$ is the equivalence dictionary of $c$.
For an AE with an original label $c_o$ but classified to be $c_t$ by the surrogate model, we call it \textit{misclassified} on the MLaaS platform if $c_o$ fails to match the response $P$ and \textit{matched} if $c_t$ matches $P$. Note that since it is possible that both the correct label and the target label are present in the response, a matched AE is not necessarily misclassified. Therefore, there is no strict relationship between the misclassification rate and the matching rate.  

Following the extended definition of misclassified and matched AEs, We further introduce two metrics for evaluating the effectiveness of transfer attacks, which is aligned to the previous studies in the lab settings. In short, transfer attack aims to find ``similar mistakes'' and what we do is to redefine what a mistake is. Note that the class mapping technique and threshold cutting are used to evaluate the transfer success rate rather than launching attacks.
\begin{definition}
  \label{def:metric_imagenet}
  We define misclassification rate to be the number of AEs that are misclassified on the MLaaS system divided by the number of AEs sent to the MLaaS system. Similarly, matching rate is defined to be the number of AEs that are matched on the MLaaS system divided by the number of AEs sent to the MLaaS system.
\end{definition}

For gender classification, every cloud has a different response format. Following
what we have done to ImageNet, we take the prediction with the higher confidence
as the final prediction. Since gender is binary, the misclassification rate and matching rate
degenerate into the same one. Thus, we further examine the Male2Female rate (M2F rate)
and Female2Male rate (F2M rate).
\begin{definition}
  \label{def:metric_adience}
  We define M2F rate to be the number of misclassified male AEs divided by the number of male AEs sent to the MLaaS system and F2M rate to be the number of misclassified female AEs divided by the number of female AEs sent to the MLaaS system.
\end{definition}

\section{Results and Analysis}

This section is divided into two parts, discussing two kinds of influencing factors respectively. The first part is about factors concerning the threat setting, \ie, platform factors, surrogate factors, adversarial algorithm factors and their joint effects. In this part, we run hierarchical ordinary least squares (OLS) regressions, on the results obtained from the ImageNet and the Adience. These regressions show how threat setting factors influence the transferability metrics, leading to a bunch of empirical observations. The reason and implication of using OLS and correlation analysis is discussed in  Appendix \ref{sec:regression} and \ref{appendix:ols_correlation}. The second part is about the factors concerning the properties of AE. In this part, we consider how the norm of adversarial perturbation, the adversarial confidence and the classification hardness affect the transferability.

The observations stated in the first part, if not explicitly claimed, are obtained from the ResNet surrogates. They  generalize if we presume the impact of surrogate architecture to transfer attacks is additively separable\footnote{For the definition of additively separation, please refer to \cite{add-separable}.} to the impact of other discussed factors.
For ease of writing, we use subscript for relations to show the $p$-value of relation tests and ``$\approx$'' to show the null hypothesis that two values are equal cannot be rejected ($p>0.1$). For example, $A<_{p=0.01}B$ means that the null hypothesis $A\ge B$ is rejected with $p$-value 0.01 and $B\approx C$ means that null hypothesis $B=C$ cannot be rejected. When we write compound inequalities, the ``$\approx$'' is treated as equivalence relation and statistical tests are done on each of them respectively. It means that $A<_{p=0.01}B\approx C$ confirms $A<_{p\le 0.01}C$ as well. In addition, we use the variable name to refer to its coefficient in the regression.

\subsection{Threat Setting Factors}
\label{sec:threat_setting_factor}

As discussed in Section \ref{sec:specific-settings}, target models on MLaaS platforms, surrogates and adversarial algorithms are the main factors in the threat setting. We use hierarchical regression to decompose the effect of these factors. Due to the large cost of adding the architecture dimension to all the experiments, we cannot run all tests on various surrogate architectures and thus the OLS regression does not include the architecture dimension in order to balance the data and avoid biased conclusions. Instead, we include a separate study of the impact of the surrogate architecture.

\begin{table*}[tbp]
  \centering
  \caption{The OLS regression result of misclassification rate and matching rate with regard to different groups of factors on the data obtained from ResNet surrogates in the image classification task. For each entry, the first number is the regression coefficient and the second number, presented in parentheses, is the corresponding standard deviation. Variables with ``is\_'' as prefix are boolean virtual variables that take the value 1 iff the corresponding conditions are true. Uninteresting values are grayed since they do not contain additional information.}
  \label{tb:imagenet_regression}
  \resizebox{0.7\linewidth}{!}{
    \begin{tabular}{ccccccccccc}
      \toprule
                                              & \multicolumn{5}{c}{\textbf{misclassification rate}} & \multicolumn{5}{c}{\textbf{matching rate}}                                                                                                                                                                             \\ \cmidrule(lr){2-6} \cmidrule(lr){7-11}
      \diagbox{IV}{Group}                     & $\mathbb{A}$                                        & $\mathbb{B}$                               & $\mathbb{C}$         & $\mathbb{D}$         & $\mathbb{E}$         & $\mathbb{F}$ & $\mathbb{G}$ & $\mathbb{H}$             & $\mathbb{I}$         & $\mathbb{J}$         \\ \midrule
      \multirow{2}{*}{is\_Google}             & 0.232\threeS                                        & 0.236\threeS                               & 0.235\threeS         & 0.190\threeS         & 0.234\threeS         & 0.100\threeS & 0.089\threeS & 0.090\threeS             & 0.090\threeS         & 0.082\threeS         \\
                                              & (0.004)                                             & (0.006)                                    & (0.005)              & (0.005)              & (0.006)              & (0.003)      & (0.004)      & (0.005)                  & (0.006)              & (0.007)              \\
      \multirow{2}{*}{is\_AWS}                & 0.071\threeS                                        & 0.075\threeS                               & 0.074\threeS         & 0.047\threeS         & 0.073\threeS         & 0.038\threeS & 0.027\threeS & 0.028\threeS             & 0.027\threeS         & 0.020\threeS         \\
                                              & (0.004)                                             & (0.006)                                    & (0.005)              & (0.005)              & (0.006)              & (0.003)      & (0.004)      & (0.005)                  & (0.006)              & (0.007)              \\
      \multirow{2}{*}{is\_Baidu}              & 0.193\threeS                                        & 0.197\threeS                               & 0.196\threeS         & 0.200\threeS         & 0.194\threeS         & 0.029\threeS & 0.019\threeS & 0.020\threeS             & 0.019\threeS         & 0.011\oneS           \\
                                              & (0.004)                                             & (0.006)                                    & (0.005)              & (0.005)              & (0.006)              & (0.003)      & (0.004)      & (0.005)                  & (0.006)              & (0.007)              \\
      \multirow{2}{*}{is\_Aliyun}             & 0.064\threeS                                        & 0.068\threeS                               & 0.067\threeS         & 0.102\threeS         & 0.065\threeS         & 0.097\threeS & 0.086\threeS & 0.087\threeS             & 0.086\threeS         & 0.078\threeS         \\
                                              & (0.004)                                             & (0.006)                                    & (0.005)              & (0.005)              & (0.006)              & (0.003)      & (0.004)      & (0.005)                  & (0.006)              & (0.007)              \\ \hdashline
      \multirow{2}{*}{is\_pretrained}         &                                                     & -0.018\threeS                              & \gray{-0.018\threeS} & \gray{-0.048\threeS} & -0.014\threeS        &              & 0.020\threeS & \gray{0.020\threeS}      & \gray{0.020\threeS}  & 0.029\threeS         \\
                                              &                                                     & (0.004)                                    & \gray{(0.002)      } & \gray{(0.005)      } & (0.005)              &              & (0.003)      & \gray{(0.003)     }      & \gray{(0.003)     }  & (0.006)              \\ \hdashline
      \multirow{2}{*}{is\_adversarial}        &                                                     & 0.010\oneS                                 & \gray{0.010\oneS   } & \gray{-0.003       } & 0.005                &              & 0.007\oneS   & \gray{0.007\twoS  }      & \gray{0.007\twoS  }  & 0.022\twoS           \\
                                              &                                                     & (0.005)                                    & \gray{(0.003)      } & \gray{(0.006)      } & (0.006)              &              & (0.004)      & \gray{(0.003)     }      & \gray{(0.003)     }  & (0.007)              \\
      \multirow{2}{*}{is\_augmented}          &                                                     & 0.004                                      & \gray{0.004        } & \gray{0.010        } & 0.014\twoS           &              & -0.006       & \gray{-0.006\oneS      } & \gray{-0.006\oneS }  & 0.003                \\
                                              &                                                     & (0.005)                                    & \gray{(0.003)      } & \gray{(0.006)      } & (0.006)              &              & (0.004)      & \gray{(0.003)     }      & \gray{(0.003)     }  & (0.007)              \\ \hdashline
      \multirow{2}{*}{is\_PGD }               &                                                     &                                            & 0.003                & \gray{0.003        } & \gray{0.003        } &              &              & 0.020\threeS             & \gray{0.020\threeS } & \gray{0.020\threeS } \\
                                              &                                                     &                                            & (0.005)              & \gray{(0.005)      } & \gray{(0.005)      } &              &              & (0.006)                  & \gray{(0.006)      } & \gray{(0.006)      } \\
      \multirow{2}{*}{is\_FGSM}               &                                                     &                                            & 0.073\threeS         & \gray{0.073\threeS } & \gray{0.073\threeS } &              &              & 0.021\threeS             & \gray{0.021\threeS } & \gray{0.021\threeS } \\
                                              &                                                     &                                            & (0.005)              & \gray{(0.005)      } & \gray{(0.005)      } &              &              & (0.006)                  & \gray{(0.006)      } & \gray{(0.006)      } \\
      \multirow{2}{*}{is\_BLB}                &                                                     &                                            & -0.048\threeS        & \gray{-0.048\threeS} & \gray{-0.048\threeS} &              &              & -0.025\threeS            & \gray{-0.025\threeS} & \gray{-0.025\threeS} \\
                                              &                                                     &                                            & (0.005)              & \gray{(0.005)      } & \gray{(0.005)      } &              &              & (0.006)                  & \gray{(0.006)      } & \gray{(0.006)      } \\
      \multirow{2}{*}{is\_CW2}                &                                                     &                                            & -0.047\threeS        & \gray{-0.047\threeS} & \gray{-0.047\threeS} &              &              & -0.026\threeS            & \gray{-0.026\threeS} & \gray{-0.026\threeS} \\
                                              &                                                     &                                            & (0.005)              & \gray{(0.005)      } & \gray{(0.005)      } &              &              & (0.006)                  & \gray{(0.006)      } & \gray{(0.006)      } \\
      \multirow{2}{*}{is\_DEEPFOOL}           &                                                     &                                            & -0.050\threeS        & \gray{-0.050\threeS} & \gray{-0.050\threeS} &              &              & 0.015\threeS             & \gray{0.015\threeS } & \gray{0.015\threeS } \\
                                              &                                                     &                                            & (0.005)              & \gray{(0.005)      } & \gray{(0.005)      } &              &              & (0.006)                  & \gray{(0.006)      } & \gray{(0.006)      } \\
      \multirow{2}{*}{is\_STEP\_LLC}          &                                                     &                                            & 0.078\threeS         & \gray{0.078\threeS } & \gray{0.078\threeS } &              &              & -0.010\oneS              & \gray{-0.010\oneS  } & \gray{-0.010\oneS  } \\
                                              &                                                     &                                            & (0.005)              & \gray{(0.005)      } & \gray{(0.005)      } &              &              & (0.006)                  & \gray{(0.006)      } & \gray{(0.006)      } \\
      \multirow{2}{*}{is\_RFGSM}              &                                                     &                                            & -0.000               & \gray{-0.000       } & \gray{-0.000       } &              &              & 0.024\threeS             & \gray{0.024\threeS } & \gray{0.024\threeS } \\
                                              &                                                     &                                            & (0.005)              & \gray{(0.005)      } & \gray{(0.005)      } &              &              & (0.006)                  & \gray{(0.006)      } & \gray{(0.006)      } \\
      \multirow{2}{*}{is\_LLC}                &                                                     &                                            & 0.002                & \gray{0.002        } & \gray{0.002        } &              &              & -0.028\threeS            & \gray{-0.028\threeS} & \gray{-0.028\threeS} \\
                                              &                                                     &                                            & (0.005)              & \gray{(0.005)      } & \gray{(0.005)      } &              &              & (0.006)                  & \gray{(0.006)      } & \gray{(0.006)      } \\ \hdashline
      \multirow{2}{*}{is\_depth\_34}          &                                                     &                                            &                      & 0.002                & 0.001                &              &              &                          & 0.002                & 0.006                \\
                                              &                                                     &                                            &                      & (0.003)              & (0.006)              &              &              &                          & (0.003)              & (0.007)              \\
      \multirow{2}{*}{is\_depth\_50}          &                                                     &                                            &                      & 0.002                & -0.003               &              &              &                          & 0.001                & 0.007                \\
                                              &                                                     &                                            &                      & (0.003)              & (0.006)              &              &              &                          & (0.003)              & (0.007)              \\ \hdashline
      \multirow{2}{*}{is\_pre$\times$adv}     &                                                     &                                            &                      &                      & -0.003               &              &              &                          &                      & -0.016\twoS          \\
                                              &                                                     &                                            &                      &                      & (0.006)              &              &              &                          &                      & (0.007)              \\
      \multirow{2}{*}{is\_pre$\times$aug}     &                                                     &                                            &                      &                      & -0.014\twoS          &              &              &                          &                      & -0.011               \\
                                              &                                                     &                                            &                      &                      & (0.006)              &              &              &                          &                      & (0.007)              \\
      \multirow{2}{*}{is\_pre$\times$depth34} &                                                     &                                            &                      &                      & -0.004               &              &              &                          &                      & 0.001                \\
                                              &                                                     &                                            &                      &                      & (0.006)              &              &              &                          &                      & (0.007)              \\
      \multirow{2}{*}{is\_pre$\times$depth50} &                                                     &                                            &                      &                      & -0.010\oneS          &              &              &                          &                      & -0.001               \\
                                              &                                                     &                                            &                      &                      & (0.006)              &              &              &                          &                      & (0.007)              \\
      \multirow{2}{*}{is\_adv$\times$depth34} &                                                     &                                            &                      &                      & 0.012\oneS           &              &              &                          &                      & -0.005               \\
                                              &                                                     &                                            &                      &                      & (0.007)              &              &              &                          &                      & (0.008)              \\
      \multirow{2}{*}{is\_adv$\times$depth50} &                                                     &                                            &                      &                      & 0.007                &              &              &                          &                      & -0.013               \\
                                              &                                                     &                                            &                      &                      & (0.007)              &              &              &                          &                      & (0.008)              \\
      \multirow{2}{*}{is\_aug$\times$depth34} &                                                     &                                            &                      &                      & -0.001               &              &              &                          &                      & -0.006               \\
                                              &                                                     &                                            &                      &                      & (0.007)              &              &              &                          &                      & (0.008)              \\
      \multirow{2}{*}{is\_aug$\times$depth50} &                                                     &                                            &                      &                      & -0.0206              &              &              &                          &                      & -0.004               \\
                                              &                                                     &                                            &                      &                      & (0.007)              &              &              &                          &                      & (0.008)              \\
      \hline
      $R^2$                                   & 0.645                                               & 0.656                                      & 0.899                & 0.899                & 0.902                & 0.382        & 0.430        & 0.587                    & 0.587                & 0.594                \\
      Adjusted $R^2$                          & 0.644                                               & 0.653                                      & 0.896                & 0.896                & 0.898                & 0.379        & 0.425        & 0.578                    & 0.577                & 0.578                \\ \midrule
      \multicolumn{9}{l}{\oneS  $p<.1$, \twoS  $p<.05$, \threeS  $p<.01$. Number of observation is 648.}                                                                                                                                                                                                                     \\
    \end{tabular}
  }
\end{table*}

\begin{table*}[]
    \centering
    \caption{The OLS regression result of F2M rate and M2F rate with regard to different groups of factors on data obtained from ResNet surrogates in the gender classification. This table is formatted similarly to Table \ref{tb:imagenet_regression}.}
    \label{tb:gender_regression}
    \resizebox{0.7\linewidth}{!}{
        \begin{tabular}{ccccccccccc}
            \toprule
                                                    & \multicolumn{5}{c}{\textbf{F2M rate}} & \multicolumn{5}{c}{\textbf{M2F rate}}                                                                                                                                                                        \\ \cmidrule(lr){2-6} \cmidrule(lr){7-11}
            \diagbox{IV}{Group}                     & $\mathbb{A}$                          & $\mathbb{B}$                          & $\mathbb{C}$         & $\mathbb{D}$         & $\mathbb{E}$        & $\mathbb{F}$ & $\mathbb{G}$  & $\mathbb{H}$         & $\mathbb{I}$         & $\mathbb{J}$        \\ \midrule

            \multirow{2}{*}{is\_AWS}                & 0.028\threeS                          & 0.023\threeS                          & -0.008               & 0.002                & 0.001               & 0.036\threeS & 0.031\threeS  & 0.002                & 0.012                & 0.005               \\
                                                    & (0.004)                               & (0.006)                               & (0.007)              & (0.007)              & (0.009)             & (0.004)      & (0.005)       & (0.007)              & (0.007)              & (0.009)             \\
            \multirow{2}{*}{is\_Baidu}              & 0.069\threeS                          & 0.064\threeS                          & 0.033\threeS         & 0.043\threeS         & 0.042\threeS        & 0.117\threeS & 0.112\threeS  & 0.083\threeS         & 0.092\threeS         & 0.086\threeS        \\
                                                    & (0.004)                               & (0.006)                               & (0.007)              & (0.007)              & (0.009)             & (0.004)      & (0.005)       & (0.007)              & (0.007)              & (0.009)             \\
            \multirow{2}{*}{is\_Aliyun}             & 0.082\threeS                          & 0.077\threeS                          & 0.047\threeS         & 0.056\threeS         & 0.055\threeS        & 0.049\threeS & 0.045\threeS  & 0.016\twoS           & 0.025\threeS         & 0.019\twoS          \\
                                                    & (0.004)                               & (0.006)                               & (0.007)              & (0.007)              & (0.009)             & (0.004)      & (0.005)       & (0.007)              & (0.007)              & (0.009)             \\ \hdashline
            \multirow{2}{*}{is\_pretrained}         &                                       & 0.019\threeS                          & \gray{0.019\threeS } & \gray{0.019\threeS } & 0.020\threeS        &              & 0.013\threeS  & \gray{0.013\threeS } & \gray{0.013\threeS } & 0.016\oneS          \\
                                                    &                                       & (0.005)                               & \gray{(0.004)      } & \gray{(0.004)      } & (0.006)             &              & (0.004)       & \gray{(0.004)    }   & \gray{(0.004)    }   & (0.008)             \\ \hdashline
            \multirow{2}{*}{is\_adversarial}        &                                       & -0.021\threeS                         & \gray{-0.021\threeS} & \gray{-0.021\threeS} & -0.033\threeS       &              & -0.016\threeS & \gray{-0.016\threeS} & \gray{-0.016\threeS} & -0.019\twoS         \\
                                                    &                                       & (0.006)                               & \gray{(0.004)      } & \gray{(0.004)      } & (0.006)             &              & (0.005)       & \gray{(0.005)    }   & \gray{(0.005)    }   & (0.009)             \\
            \multirow{2}{*}{is\_augmented}          &                                       & 0.008                                 & \gray{0.008\oneS   } & \gray{0.008\oneS   } & 0.014\oneS          &              & 0.010\oneS    & \gray{0.010\oneS  }  & \gray{0.010\twoS  }  & 0.011               \\
                                                    &                                       & (0.006)                               & \gray{(0.004)      } & \gray{(0.004)      } & (0.006)             &              & (0.005)       & \gray{(0.005)    }   & \gray{(0.005)    }   & (0.009)             \\ \hdashline
            \multirow{2}{*}{is\_PGD }               &                                       &                                       & 0.056\threeS         & \gray{0.056\threeS}  & \gray{0.056\threeS} &              &               & 0.047\threeS         & \gray{0.047\threeS}  & \gray{0.047\threeS} \\
                                                    &                                       &                                       & (0.008)              & \gray{(0.008)     }  & \gray{(0.007)     } &              &               & (0.008)              & \gray{(0.008)     }  & \gray{(0.008)     } \\
            \multirow{2}{*}{is\_FGSM}               &                                       &                                       & 0.092\threeS         & \gray{0.092\threeS}  & \gray{0.092\threeS} &              &               & 0.070\threeS         & \gray{0.070\threeS}  & \gray{0.070\threeS} \\
                                                    &                                       &                                       & (0.008)              & \gray{(0.008)     }  & \gray{(0.007)     } &              &               & (0.008)              & \gray{(0.008)     }  & \gray{(0.008)     } \\
            \multirow{2}{*}{is\_BLB}                &                                       &                                       & -0.001               & \gray{-0.001      }  & \gray{-0.001      } &              &               & 0.015\oneS           & \gray{0.015\oneS  }  & \gray{0.015\oneS  } \\
                                                    &                                       &                                       & (0.008)              & \gray{(0.008)     }  & \gray{(0.007)     } &              &               & (0.008)              & \gray{(0.008)     }  & \gray{(0.008)     } \\
            \multirow{2}{*}{is\_CW2}                &                                       &                                       & 0.001                & \gray{0.001       }  & \gray{0.001       } &              &               & 0.014\oneS           & \gray{0.014\oneS  }  & \gray{0.014\oneS  } \\
                                                    &                                       &                                       & (0.008)              & \gray{(0.008)     }  & \gray{(0.007)     } &              &               & (0.008)              & \gray{(0.008)     }  & \gray{(0.008)     } \\
            \multirow{2}{*}{is\_DEEPFOOL}           &                                       &                                       & -0.013               & \gray{-0.013      }  & \gray{-0.013\oneS } &              &               & 0.008                & \gray{0.008       }  & \gray{0.008       } \\
                                                    &                                       &                                       & (0.008)              & \gray{(0.008)     }  & \gray{(0.007)     } &              &               & (0.008)              & \gray{(0.008)     }  & \gray{(0.008)     } \\
            \multirow{2}{*}{is\_STEP\_LLC}          &                                       &                                       & 0.070\threeS         & \gray{0.070\threeS}  & \gray{0.070\threeS} &              &               & 0.052\threeS         & \gray{0.052\threeS}  & \gray{0.052\threeS} \\
                                                    &                                       &                                       & (0.008)              & \gray{(0.008)     }  & \gray{(0.007)     } &              &               & (0.008)              & \gray{(0.008)     }  & \gray{(0.008)     } \\
            \multirow{2}{*}{is\_RFGSM}              &                                       &                                       & 0.039\threeS         & \gray{0.039\threeS}  & \gray{0.039\threeS} &              &               & 0.034\threeS         & \gray{0.034\threeS}  & \gray{0.034\threeS} \\
                                                    &                                       &                                       & (0.008)              & \gray{(0.008)     }  & \gray{(0.007)     } &              &               & (0.008)              & \gray{(0.008)     }  & \gray{(0.008)     } \\
            \multirow{2}{*}{is\_LLC}                &                                       &                                       & 0.030\threeS         & \gray{0.030\threeS}  & \gray{0.030\threeS} &              &               & 0.022\threeS         & \gray{0.022\threeS}  & \gray{0.022\threeS} \\
                                                    &                                       &                                       & (0.008)              & \gray{(0.008)     }  & \gray{(0.007)     } &              &               & (0.008)              & \gray{(0.008)     }  & \gray{(0.008)     } \\ \hdashline
            \multirow{2}{*}{is\_depth\_34}          &                                       &                                       &                      & -0.009\twoS          & 0.000               &              &               &                      & -0.008\oneS          & 0.008               \\
                                                    &                                       &                                       &                      & (0.004)              & (0.009)             &              &               &                      & (0.005)              & (0.009)             \\
            \multirow{2}{*}{is\_depth\_50}          &                                       &                                       &                      & -0.019\threeS        & -0.023\threeS       &              &               &                      & -0.020\threeS        & -0.001              \\
                                                    &                                       &                                       &                      & (0.004)              & (0.009)             &              &               &                      & (0.005)              & (0.009)             \\ \hdashline
            \multirow{2}{*}{is\_pre$\times$adv}     &                                       &                                       &                      &                      & -0.002              &              &               &                      &                      & 0.005               \\
                                                    &                                       &                                       &                      &                      & (0.009)             &              &               &                      &                      & (0.009)             \\
            \multirow{2}{*}{is\_pre$\times$aug}     &                                       &                                       &                      &                      & 0.010               &              &               &                      &                      & 0.027\threeS        \\
                                                    &                                       &                                       &                      &                      & (0.009)             &              &               &                      &                      & (0.009)             \\
            \multirow{2}{*}{is\_pre$\times$depth34} &                                       &                                       &                      &                      & -0.003              &              &               &                      &                      & -0.014              \\
                                                    &                                       &                                       &                      &                      & (0.009)             &              &               &                      &                      & (0.009)             \\
            \multirow{2}{*}{is\_pre$\times$depth50} &                                       &                                       &                      &                      & -0.009              &              &               &                      &                      & -0.026\threeS       \\
                                                    &                                       &                                       &                      &                      & (0.009)             &              &               &                      &                      & (0.009)             \\
            \multirow{2}{*}{is\_adv$\times$depth34} &                                       &                                       &                      &                      & 0.002               &              &               &                      &                      & -0.004              \\
                                                    &                                       &                                       &                      &                      & (0.010)             &              &               &                      &                      & (0.011)             \\
            \multirow{2}{*}{is\_adv$\times$depth50} &                                       &                                       &                      &                      & 0.035\threeS        &              &               &                      &                      & 0.006               \\
                                                    &                                       &                                       &                      &                      & (0.010)             &              &               &                      &                      & (0.011)             \\
            \multirow{2}{*}{is\_aug$\times$depth34} &                                       &                                       &                      &                      & -0.026\twoS         &              &               &                      &                      & -0.022\twoS         \\
                                                    &                                       &                                       &                      &                      & (0.010)             &              &               &                      &                      & (0.011)             \\
            \multirow{2}{*}{is\_aug$\times$depth50} &                                       &                                       &                      &                      & -0.010              &              &               &                      &                      & -0.022\twoS         \\
                                                    &                                       &                                       &                      &                      & (0.010)             &              &               &                      &                      & (0.011)             \\
            \hline
            $R^2$                                   & 0.150                                 & 0.218                                 & 0.558                & 0.575                & 0.600               & 0.352        & 0.395         & 0.531                & 0.550                & 0.575               \\
            Adjusted $R^2$                          & 0.147                                 & 0.210                                 & 0.546                & 0.562                & 0.580               & 0.350        & 0.389         & 0.518                & 0.535                & 0.554               \\ \midrule
            \multicolumn{11}{l}{\oneS  $p<.1$, \twoS  $p<.05$, \threeS  $p<.01$. Number of observation is 486.}                                                                                                                                                                                            \\
        \end{tabular}
    }
\end{table*}

Table \ref{tb:imagenet_regression} shows the OLS regression result on the ImageNet dataset. Each column (regression group) is a different OLS regression with different independent variable (IV) group. For each column, variables left blank are not included in the regression. The table follows a standard representation of hierarchical regression and readers who are unfamiliar with this representation can find more explanations on how to read this table in Appendix \ref{appendix:ols_explanation}. Detailed codes for conducting the OLS analysis is included in the released code repository. Regression $\mathbb{A}$ and $\mathbb{F}$ decompose the effect of target models, revealing how well different target MLaaS platforms behave to defend transfer attacks. Regression $\mathbb{B}$ and $\mathbb{G}$ further decompose the effect of pretraining and data enrichment factors, intending to reveal how surrogate training affects the transferability. Adversarial algorithm factors' effect are further decomposed in regression $\mathbb{C}$ and $\mathbb{H}$. In regression $\mathbb{D}$ and $\mathbb{I}$, surrogate depth's effect is decomposed as well. In the result of these regressions, data enrichment and surrogate depth are not clearly correlated to transfer attack, thus regression $\mathbb{E}$ and $\mathbb{J}$ are designed to reveal if there are joint effects between surrogate model setting and training techniques. Table \ref{tb:gender_regression} shows the result on the Adience dataset, and is organized similarly.

In the regression, multicollinearity exists among platform factors, adversarial algorithm factors and surrogate depth factors. For example, one and only one of the platform factors is 1, which makes their coefficients not unique in the regression. To avoid this problem, we choose a baseline setting: UAP attack for adversarial algorithm factors and ResNet-18 for surrogate depth factors. This means that for the settings with UAP attack, all indicator variables for the adversarial algorithms are zero. Similarly, for the settings with ResNet-18 surrogate, all indicators for the surrogate depth are zero. No baseline platform is set because the constant term of the regression can be merged to the platform factors.

\subsubsection{Platform Factors}

Platform factors represent the vulnerability of the target model. A larger coefficient for a platform means that transfer attacks are more likely to succeed on this platform. However, as we use a restricted range of classes  and a manually designed class mapping to decide whether a transfer attack is successful (see Section \ref{sec:metric} for details), the following discussions do not provide any guarantee nor comparison for the robustness of these platforms when different settings are applied. We are careful not to generalize our conclusions, and readers are discouraged to compare the MLaaS systems by their performance in these specific tasks and datasets.

Regression \reg{A} and Regression \reg{F} decomposes the platform factors in Table \ref{tb:imagenet_regression} and Table \ref{tb:gender_regression}, respectively. Since these two regressions run on all settings but only include platform factors in the IV group, the coefficient of the platform factors represents an average of metric values over all settings. In other words, if attackers pick their settings uniformly at random, then they are expecting the coefficient as their metric value. This provides us with some knowledge about ``inexperienced attackers'' and is a good representation of the robustness of the target model.

By comparing the coefficients, we get the following relations (lower is better):
\begin{enumerate}[(i)]
  \item In the object classification task, Aliyun $\approx$ AWS $<_{p<0.001}$ Baidu $<_{p<0.001}$ Google on the misclassification rate, and Baidu $<_{p=0.068}$ AWS $<_{p<0.001}$ Aliyun $\approx$ Google on the matching rate.
  \item In the gender classification task, AWS $<_{p<0.001}$ Baidu $<_{p=0.032}$ Aliyun on the F2M rate, and AWS $<_{p=0.013}$ Aliyun $<_{p<0.001}$ Baidu on the M2F rate.
\end{enumerate}

It shows that although these platforms have similar accuracies on the clean dataset (see Figure \ref{fig:threshold-accuracy} for details), their robustness against transfer attack varies. In particular, the ranking of the robustness is not the same to the ranking of their accuracy. This result suggests that model accuracy does not necessarily guarantee robustness against transfer attack, even in the real applications. In addition, no single platform has superior robustness in different kinds of transfer attacks (untargeted vs targeted, F2M vs M2F), although AWS is the best in the gender classification. In particular, when compared with other platforms, Aliyun is good at defending the untargeted attacks but not the targeted attacks. On the contrary, Baidu is good at defending the targeted attcks but not the untargeted attacks. This means that a model's robustness cannot be naively measured by its robustness against specific kind of attacks. Therefore, since the coefficients are significantly positive, some even exceeding 0.2, the platforms should take transfer attack more seriously due to their ignorable marginal cost because even inexperienced attackers can get over 20\% work done with a very small cost.

Furthermore, we observe that the matching rate is significantly positive for all the platforms. This is different to the conclusion of Liu \etal \cite{liu2016delving} that targeted transfer attack almost never transfer, by showing that targeted transfer attack can succeed in the real applications.

\begin{observation} \label{obs:platform}
  In the real transfer attack, the difficulty of attacking a target model is not directly related to its accuracy, \ie, a target with higher accuracy is possible to be more vulnerable to transfer attacks. No single platform has superior robustness in different kinds of transfer attacks (untargeted vs targeted, F2M vs M2F). Therefore, the threat of transfer attacks in the real applications should be treated seriously because it has a non-trivial success rate and a low cost.
\end{observation}

\subsubsection{Pretraining and Surrogate Dataset Factors}

Arguably, pretraining plays an important role in training a decent model with few efforts, thus is attempting to be applied to train surrogate models. Regression \reg{B} and Regression \reg{G} decompose the effect of pretraining in Table \ref{tb:imagenet_regression} and Table \ref{tb:gender_regression}, respectively. By looking at its coefficients, we get the following relations:
\begin{enumerate}[(i)]
  \item In the object classification task, pretraining has a significantly negative impact on the misclassification rate but a significantly positive effect on the matching rate.
  \item In the gender classification task, pretraining has a significantly positive effect on both the F2M rate and the M2F rate.
\end{enumerate}

The observation in the object classification task is intuitively confusing.
In the research common sense, pretrained surrogates should be more similar to the target model because they both have a low error rate. Therefore, better similarity should guarantee better transferability of the crafted AEs. The empirical result, on the contrary, tells us that pretrained surrogates are worse choices for untargeted transfer attack but better for targeted transfer attack. The result is unlikely  caused by the potential joint effects because further decomposition shown in Table \ref{tb:imagenet_regression} supports it as well. In addition, we check the surrogate accuracy and confirm that pretrained surrogates have a roughly 5\% improvement on accuracy for all model types.

This finding shows that the ``surrogate similarity'' concept is not suitable, at least not straightforward, to describe transfer attacks. On the one hand, suppose pretraining does improve the similarity between the surrogate and target, then both the misclassification rate and matching rate should be improved because better similarity ensures better transferability. On the other hand, suppose pretraining reduces the similarity, then both metrics should be lowered. However, neither can explain this result. Therefore, if we want to define ``surrogate similarity'', we should not define it via the accuracy gap.

In the gender classification task, however, pretraining improves both metrics. This suggests that a simpler task may benefit more from pretraining. Although we are not aware of any existing result that can explain this, we have an intuitive hypothesis about why pretraining works in simple tasks. Bahri \etal \cite{doi:10.1146/annurev-conmatphys-031119-050745} mentioned a result that in the random Gaussian landscape, the number of local optima increases exponentially when the loss decreases. If we generalize this result and hypothesize that the number of local optima increases exponentially when the task complexity increases, then the empirical finding follows. This is because under this hypothesis, the ten-class object classification task is more complex than the binary gender classification task and thus much more local optima is present. Therefore, although pretraining improves the model's accuracy, it leads the surrogate into a local optima that is different to the local optima in which the target model locates, which makes the boundary of the surrogate and the target model less similar. The targeted transfer attack, however, relies less on the boundary. Instead, it relies more on the correct direction of the perturbation towards the target class, and a similar direction towards the target class is enforced by performing better on the task. This rationale may explain why the matching rate is improved while the misclassification rate is decreased by pretraining.

Regression \reg{B} and Regression \reg{G} decompose the effect of surrogate dataset factors as well. However, adversarial training and data augmentation do not show strong correlation to the transfer attack. Therefore, we leave the discussion about these factors to the later section which considers the joint effect of surrogate-level factors.

\begin{observation} \label{obs:pre}
  Pretraining improves targeted transfer attacks but not untargeted transfer attacks. This implies an appropriate definition of model similarity is extremely difficult.
\end{observation}

\subsubsection{Adversarial Algorithm Factors} \label{subsec:adv_algo}

Since the finding of the adversarial property of neural networks, adversarial algorithm is the main object being studied, making it relatively well-understood among all components of a transfer attack. However, their transferability in the real attack scenario is under-explored.

Regression \reg{C} and Regression \reg{H} decompose the effect of adversarial algorithms in Table \ref{tb:imagenet_regression} and Table \ref{tb:gender_regression}, respectively. By comparing its coefficients, we get the following relations:
\begin{enumerate}[(i)]
  \item In the object classification task, DeepFool $\approx$ BLB $\approx$ CW2 $<_{p<0.001}$ LLC $\approx$ RFGSM $\approx$ UAP $\approx$ PGD $<_{p<0.001}$ FGSM $\approx$ Step-LLC on the misclassification rate, and LLC $\approx$ CW2 $\approx$ BLB $<_{p=0.008}$ Step-LLC $<_{p=0.068}$ UAP $<_{p=0.008}$ DeepFool $\approx$ PGD $\approx$ FGSM $\approx$ RFGSM.
  \item In the gender classification task, DeepFool $\approx$ BLB $\approx$ UAP $\approx$ CW2 $<_{p<0.001}$ LLC $\approx$ RFGSM $<_{p=0.031}$ PGD $<_{p=0.062}$ Step-LLC $<_{p=0.004}$ FGSM on the F2M rate. For the M2F rate, we have UAP $\approx$ DeepFool, CW2 $\approx$ BLB $\approx$ LLC and RFGSM $<_{p=0.092}$ PGD $\approx$ Step-LLC $<_{p=0.021}$ FGSM.
\end{enumerate}

Two patterns are especially significant. First, the strong adversarial algorithms (BLB and CW2) transfer worse than most of the algorithms on all the metrics, which is unexpected. In particular, although they are targeted, their matching rates are lower than untargeted algorithms. In addition, FGSM performs well in all kind of transfer attacks, although it uses less information from the surrogate. Second, single-step algorithms have a higher misclassification rate than their iterative counterparts (Step-LLC $>$ LLC and FGSM $>$ PGD). Su \etal \cite{su2018robustness} confirms that FGSM transfers better than PGD as well. This result is surprising because iterative algorithms are more powerful and exploit more information from the surrogate model. Combining these two facts, we can see that attacks that use too much information of the surrogate are less likely to transfer, and the probably most transferable information is the gradient with regard to the seed image.

\begin{observation} \label{obs:algorithm}
  In the real applications, strong adversarial algorithms, \eg, CW2, might have weak transferability. In addition, single-step algorithms transfer better than iterative algorithms, \eg, FGSM $>$ PGD. This suggests the probably most transferable information is the gradient with regard to the seed image.
\end{observation}

\subsubsection{Surrogate Depth Factors}

Choosing the surrogate depth is important for a good surrogate model. Demontis \etal \cite{demontis2018adversarial} pointed out that simple surrogates are better than complex surrogates. However, in practice, this brings up a question: a too simple surrogate cannot perform reasonably on the task. For example, the most complex surrogate that Demontis \etal applied was a two-layer neural network. However, this particular architecture is too simple to be the surrogate in our tasks, getting roughly 15\% accuracy on the test data for the object classification task. Therefore, the impact of surrogate depth in the real scenario is still under-explored.

Regression \reg{D} and \reg{I} decompose the effect of surrogate depth in Table \ref{tb:imagenet_regression} and Table \ref{tb:gender_regression}, respectively. By comparing the coefficients, we can see that for the gender classification, ResNet-18 indeed performs better than ResNet-34 and ResNet-50. However, for the object classification task, ResNet-18, ResNet-34 and ResNet-50 essentially have the same performance. This shows simpler surrogates do not necessarily have better transferability in the real transfer attack, especially for complex tasks. We further conduct two experiments for the object classification task. One of them aims to show that on local targets, an appropriate ResNet surrogate preserves better transferability than both simpler and deeper surrogates. The other aims to show that for VGG surrogates, the same phenomenon is observed as well.

\begin{enumerate}[(i)]
  \item \textbf{Experiments on Local Targets}

        To make our settings more similar to the previous study by Demontis \etal \cite{demontis2018adversarial}, we apply AEs crafted from ResNet surrogates with different depths against a local VGG-16 target model. We calculate the difference in misclassification rate among various depths for each threat setting which consists of different pretraining factors and attack algorithm factors. The surrogate dataset is fixed to raw, \ie, no data enrichment is used. Figure \ref{fig:local_vgg_target} shows the distribution of the difference between the ResNet-34 and the ResNet-18 surrogates and the difference between the ResNet-50 and the ResNet-34 surrogates.

        \begin{figure}
          \centering
          \includegraphics[width=0.4\linewidth]{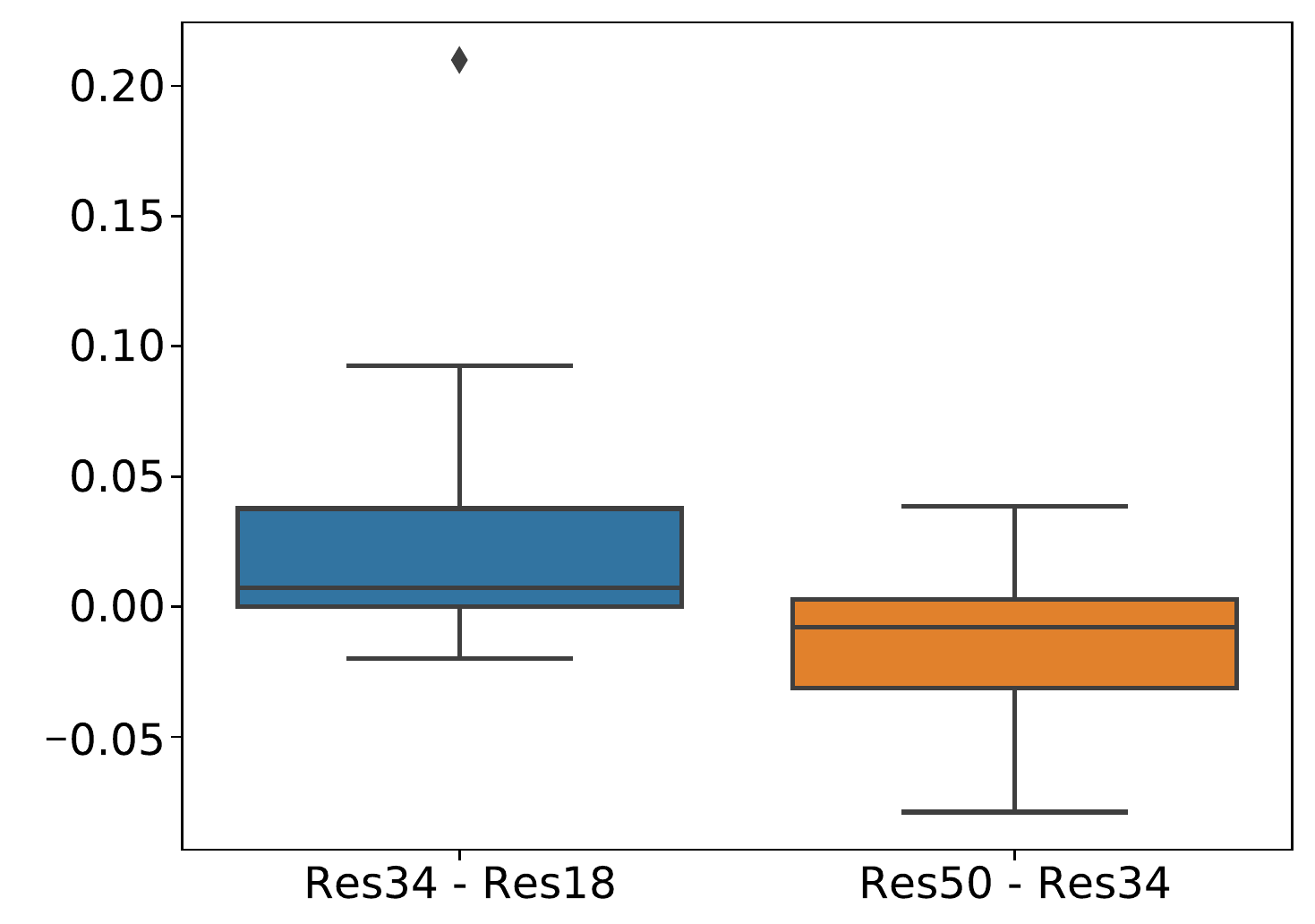}
          \caption{The box plot of the difference in misclassification rate against the local VGG target. The diamond represents an outlier.}
          \label{fig:local_vgg_target}
        \end{figure}

        It can be seen from Figure \ref{fig:local_vgg_target} that the ResNet-34 surrogates have higher transferability than ResNet-18 but the ResNet-50 surrogates have lower transferability than ResNet-34. By performing Wilcoxon test\cite{Wilcoxon}, we get that the first conclusion has $p$-value 0.028 and the second has $p$-value 0.035, which are statistically sufficient. This result agrees with our hypothesis that an appropriate surrogate complexity is better than lower and higher complexity.

  \item \textbf{Experiments on VGG Surrogates}

        To make sure that this phenomenon is not restricted to ResNet surrogates, we extend the result to VGG surrogates. We use VGG-11, VGG-13, VGG-16 and VGG-19 as surrogates and evaluate the transferability of the crafted AEs on the MLaaS systems. These surrogates are obtained from pretrained models and the raw dataset is used to fine-tune them, \ie, \emph{is\_pretrained} is fixed to True and \emph{is\_augmented} and \emph{is\_adversarial} are fixed to False.

        \begin{table}[]
          \caption{The OLS results on VGG surrogates.}
          \label{tb:vgg_surrogate}
          \centering
          \resizebox{0.5\linewidth}{!}{
            \begin{tabular}{cccc}
              \multicolumn{1}{l}{} & coef             & std err & $p$-value \\ \toprule
              is\_google           & 0.1711           & 0.018   & 0.000     \\
              is\_aws              & 0.0101           & 0.018   & 0.570     \\
              is\_baidu            & 0.1444           & 0.018   & 0.000     \\
              is\_aliyun           & 0.0240           & 0.018   & 0.177     \\ \midrule
              is\_PGD              & 0.0482           & 0.019   & 0.014     \\
              is\_FGSM             & 0.2016           & 0.019   & 0.000     \\
              is\_BLB              & -0.0006          & 0.019   & 0.977     \\
              is\_CW2              & -0.0157          & 0.019   & 0.421     \\
              is\_DEEPFOOL         & -0.0207          & 0.019   & 0.288     \\
              is\_STEP\_LLC        & 0.1575           & 0.019   & 0.000     \\
              is\_RFGSM            & 0.0749           & 0.019   & 0.000     \\
              is\_LLC              & 0.0244           & 0.019   & 0.210     \\ \midrule
              \textbf{is\_13}      & \textbf{0.0006 } & 0.013   & 0.960     \\
              \textbf{is\_16}      & \textbf{0.0780 } & 0.013   & 0.000     \\
              \textbf{is\_19}      & \textbf{-0.0003} & 0.013   & 0.983     \\ \bottomrule
            \end{tabular}
          }
        \end{table}

        Table \ref{tb:vgg_surrogate} shows the result of OLS regression. The emphasized numbers in the table show that using VGG-13, 16 and 19 has a 0.0006, 0.078 and -0.0003 improvement in the misclassification rate respectively when compared to VGG-11. By looking at the $p$-values, we observe that VGG-11 $\approx$ VGG-13 $\approx$ VGG-19 $<_{p<0.001}$ VGG-16, making VGG-16 a better surrogate than its simpler and deeper versions. This result shows that the phenomenon that a surrogate with appropriate depth is better than other surrogates is not restricted to ResNet surrogates but holds for VGG surrogates as well.
\end{enumerate}

\begin{observation}
  Surrogate complexity, defined by the depth of the surrogate, has a non-monotonic effect on the transferability. A surrogate with appropriate depth is better than both simpler and deeper surrogates.
\end{observation}

\subsubsection{Joint Effect of Surrogate-Level Factors}

Various factors are included to train a surrogate model, \ie, surrogate dataset, pretraining and the choice of surrogate depth. Therefore, their interactions are of particular interest, since these factors affect the attack via the surrogate as a whole.

Regression \reg{E} and Regression \reg{J} decompose the joint effect of surrogate-level factors in Table \ref{tb:imagenet_regression} and Table \ref{tb:gender_regression}, respectively. Before we continue, some interpretation of regression $\mathbb{E}$ and $\mathbb{J}$ should be explained to avoid misunderstanding of the result.
First, the indicator variable of the decomposed factor have a different statistical interpretation when compared to other regressions due to the joint terms included.
For example, in regression $\mathbb{E}$ and $\mathbb{J}$, the coefficient of \emph{is\_pretrained} represents the influence of pretraining when ResNet-18 surrogate is used without data enrichment. This is because the joint effect is characterized by the joint terms while previously the coefficient is not conditioned on the choice of ResNet-18 and the absence of data enrichment.
Similar rules apply to the coefficients of \emph{is\_adversarial}, \emph{is\_augmented} and surrogate depth factors.
Second, a significantly positive coefficient of the joint term does not necessarily mean that using them jointly is good for transfer attack, vice versa.
This is because the joint terms should be added with the individual terms to characterize the joint effect. For example, for the Regression \reg{E} shown in Table \ref{tb:imagenet_regression}, the coefficient of \emph{is\_pre$\times$adv} is -0.003. To recover the effect of the joint usage of pretraining and adversarial training, we should add this term with \emph{is\_pretraining} and \emph{is\_adversarial}, which is (-0.003) + (-0.014) + (+0.005) = -0.012.
Third, we cannot conclude that two factors are approximately independent, even though the joint term is not significant. Instead, this should be digested as that the joint effect is not significant enough to be observed.

We only discuss joint terms with a significantly non-zero coefficient. The first interesting phenomenon is that almost no joint term has consistent effects on different kinds of transfer attack, \ie, if they show significant improvement on untargeted transfer attack, then they almost always do not show significant improvement on targeted transfer attack.  This phenomenon further supports Observation \ref{obs:pre} by showing that pretraining is not the only one that has this feature. Second, the interaction between the surrogate-level factors is extremely complex, as no common effect is observed in both the Table \ref{tb:imagenet_regression} and Table \ref{tb:gender_regression}. For example, in Table \ref{tb:imagenet_regression} we observe significantly negative coefficient for \emph{is\_pre$\times$aug} on the misclassification rate. However, in Table \ref{tb:gender_regression} this coefficient becomes insignificantly positive. Therefore, choosing a good surrogate model actually needs trial and error, and is highly task-specific.

\begin{observation}
  The interaction between surrogate-level factors is highly chaotic and task-specific. Training a good surrogate needs trial and error.
\end{observation}

\subsubsection{Importance of Different Factors}

We have discussed the effect of many factors, but we still want to know which factor contributes the most to the transferability of AEs. The contribution of factors can be measured by the changes of $R^2$ when the factors are included.\footnote{$R^2$ is a  statistic that measures the ratio of model's explained variance to the total variance. When additional independent variables are included in a linear regression, $\Delta R^2$ measures the contribution of additional variables.}

By comparing the $\Delta R^2$ of hierarchical regressions in Table \ref{tb:imagenet_regression} and Table \ref{tb:gender_regression}, we can see the most important factors are the platform factors and the adversarial algorithm factors, with a $\Delta R^2$ ranging from 0.150 to 0.645 and from 0.136 to 0.340, respectively. This means that attackers benefit the most from setting a good target platform and a good adversarial algorithm. While the target platform is predetermined, the best improvement for the attack is to choose an appropriate attack algorithm, e.g., FGSM.

\begin{observation}
  Among all the discussed factors, the easiest and most beneficial practice to improve a transfer attack is to apply an appropriate adversarial algorithm, \eg, FGSM.
\end{observation}

\subsubsection{Surrogate Architecture Factors}

\begin{figure}
  \centering
  \includegraphics[width=.8\linewidth]{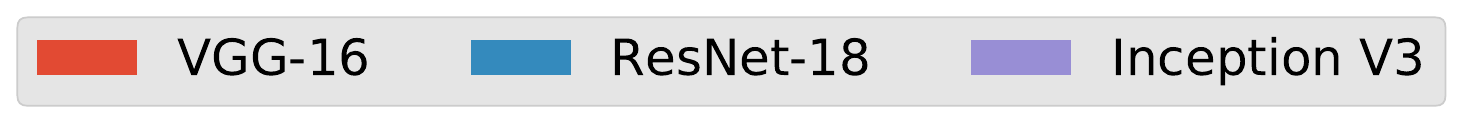}

  \includegraphics[width=.9\linewidth]{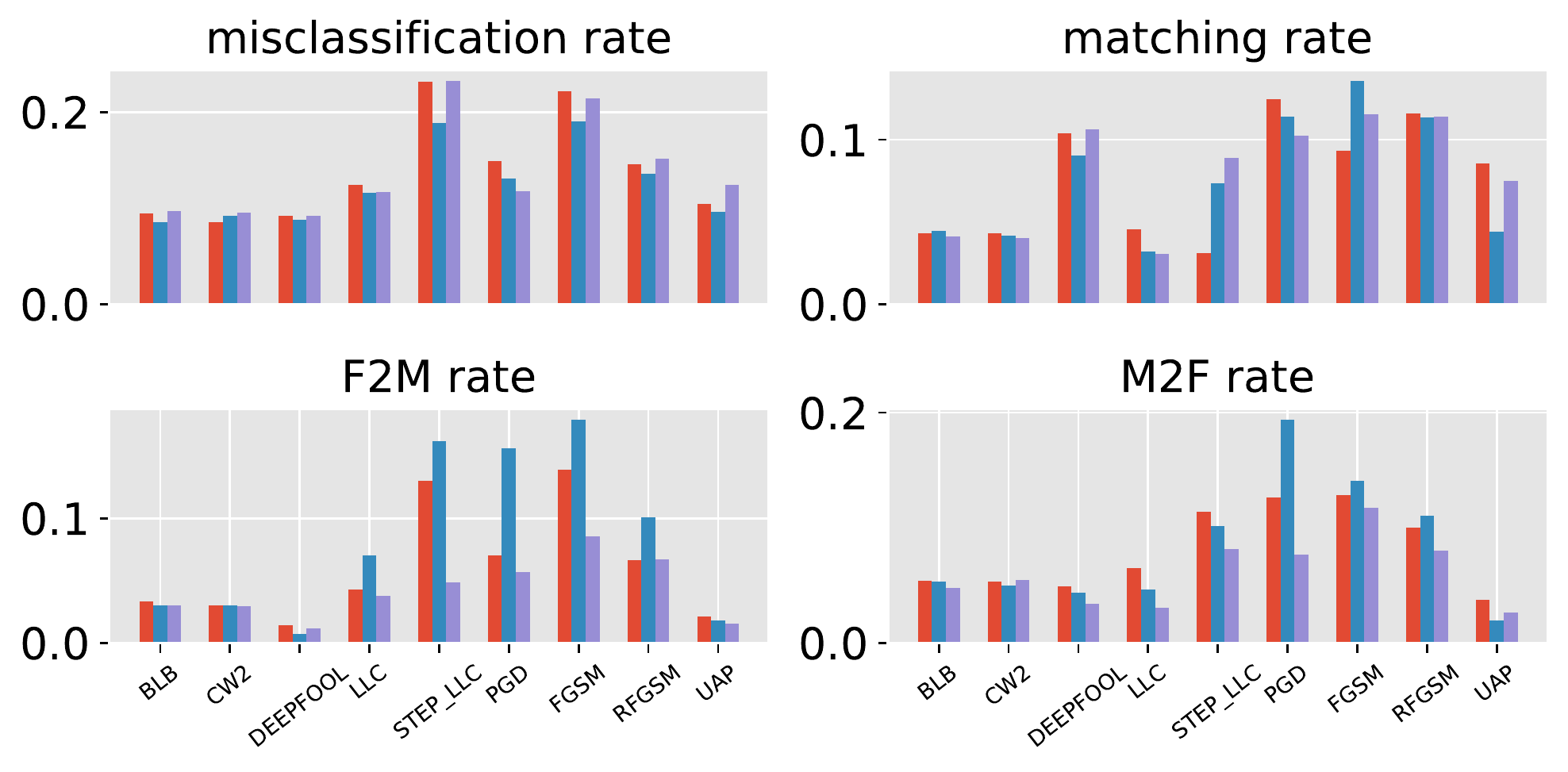}
  \caption{The transfer attack results grouped by surrogate architectures and attack algorithms. The result of each bar is averaged on the four platforms.}
  \label{fig:architecture}
\end{figure}

We have completed the OLS analysis of transfer attack in the image classification and the gender classification, leaving an important factor behind: the impact of surrogate architecture. Figure \ref{fig:architecture} plots metrics against adversarial algorithms grouped by surrogate architectures. Unlike the conclusion from Su \etal \cite{su2018robustness} that AEs crafted from most surrogates only transfer in their own surrogate family except VGG,  we can see from Figure \ref{fig:architecture} that no dominant architecture exists in the real scenario. The result suggests surrogate families other than VGG worth attention in the real transfer attack as well.

\begin{observation}
  No dominant architecture family is found in the real transfer attack.
  \label{observe:arch}
\end{observation}

\subsection{Sample Property Factors}
\label{sec:sample}

In this section, we dive deeper into the correlations between
transferability and sample-level properties.
Specifically, we study the following three properties: the norm of adversarial perturbation,
the adversarial confidence (\eg, $\kappa$ in CW attack\cite{carlini2016evaluating}) and the intrinsic classification hardness of seed images.

\subsubsection{Connection Between Transferability and the Norm of the Adversarial Perturbation}
\label{sec:norm}
\label{sec:correlation}

\begin{figure}
  \centering
  \begin{subfigure}{.37\linewidth}
    \centering
    \includegraphics[width=\textwidth]{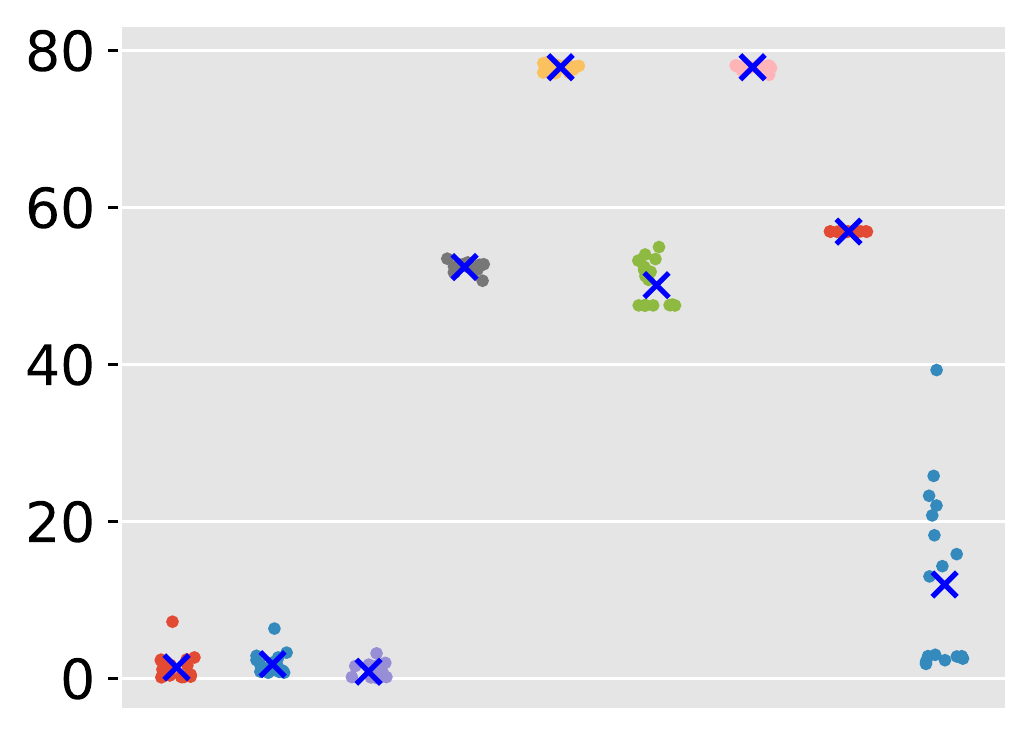}
    \caption{ $L_2$ norm}
    \label{fig:L2}
  \end{subfigure}
  \begin{subfigure}{.37\linewidth}
    \centering
    \includegraphics[width=\textwidth]{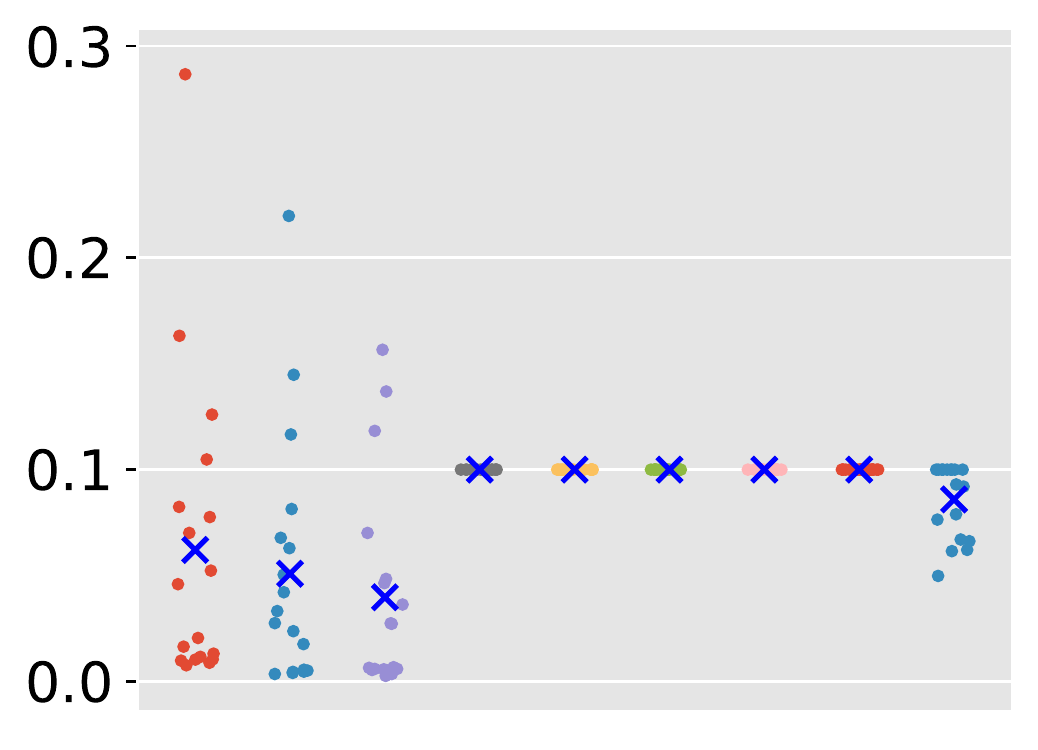}
    \caption{ $L_\infty$ norm}
    \label{fig:LINF}
  \end{subfigure}
  \begin{subfigure}{.2\linewidth}
    \centering
    \includegraphics[width=40pt, height=60pt]{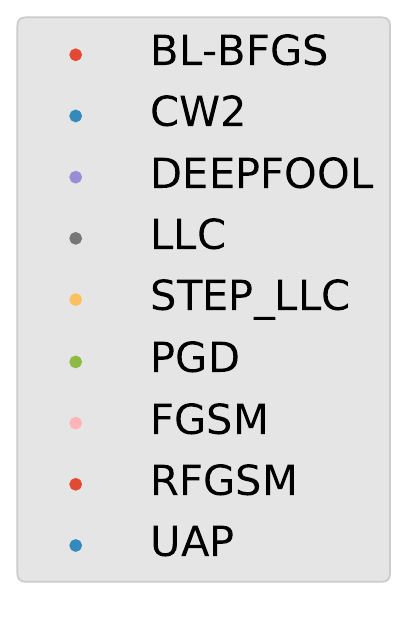}
  \end{subfigure}
  \caption{The $L_2$ and $L_{\infty}$ norm of perturbation for AEs.}
  \label{fig:norm}
\end{figure}

So far, relaxing the perturbation norm budget of adversarial
attacks has been believed to be a general method to increase transferability. We find this conclusion still holds in the real setting, hence we aim to provide an answer for a deeper question: does the type of norm matter in deciding transferability?

\begin{figure}
  \centering
  \includegraphics[width=.7\linewidth]{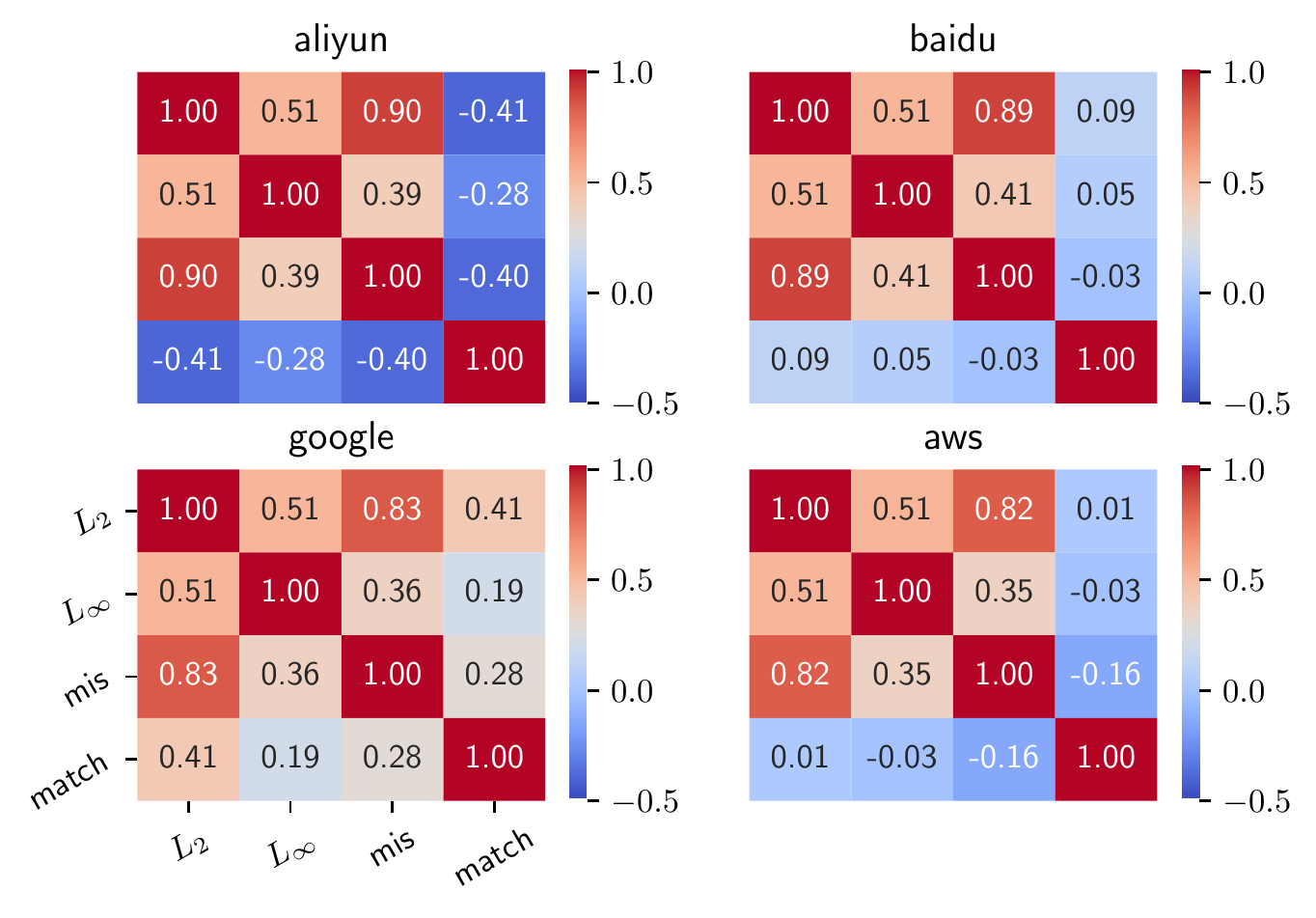}
  \caption{The correlation of norm and transferability.}
  \label{fig:correlation}
\end{figure}

Figure \ref{fig:norm} plots the average norm of the adversarial perturbations generated by each attack
algorithm in the image classification task. It is interesting to notice that the order of adversarial algorithms in terms of $L_2$ norm in Figure \ref{fig:L2} is roughly aligned to the order in terms of misclassification rate shown in Section \ref{subsec:adv_algo}. Figure \ref{fig:correlation} further computes the correlation matrix between transferability metrics and norms, from which we observe the correlation between $L_2$ norm and misclassification rate is extremely high, exceeding 0.8. The correlation between $L_{\infty}$ and misclassification rate is far weaker, up to 0.41. In fact, we prove in Appendix \ref{sec:infty-mis} that even this correlation is largely due to the dependence of $L_\infty$ on $L_2$ norm. This finding motivates us to hypothesize that a perturbation with larger $L_2$ norm and fixed $L_\infty$ norm improves transferability. As a validation of this hypothesis, we repeat to generate perturbations sampled from set $\{-0.1,0.1\}^d$ randomly and stop until either an adversarial perturbation that deceives the surrogate is found or a maximum iteration, 1000 times, is reached. In this way, we get adversarial perturbation with the largest $L_2$ norm while keeping the $L_\infty$ norm fixed at 0.1. These AEs, though unaware of the explicit gradient information, achieve an average misclassification rate of 39.1\% on Google, 16.5\% on AWS, 25.4\% on Aliyun and 36.7\% on Baidu, respectively, which is higher than many attack algorithms with the same $L_\infty$ norm budget. On the
contrary, AEs with large $L_{\infty}$ norm of perturbations but small $L_2$
norm, such as those generated by BLB and CW, only have trivial
transferability as shown in Section \ref{subsec:adv_algo}. Therefore, we conclude that transfer attacks are closely related to the $L_2$ norm of the adversarial perturbation but not the $L_\infty$ norm, although human vision systems are believed to be insensitive to $L_2$ norms \cite{DBLP:journals/tci/ZhaoGFK17}.

\begin{observation}
  The transferability of the AEs is very closely related to the $L_2$ norm of the perturbation but not the $L_\infty$ norm. Increasing the $L_2$ norm while keeping $L_\infty$ norm fixed can increase the transferability of the AEs.
\end{observation}



\subsubsection{Connection between Transferability and the Adversarial Confidence}
\label{sec:kappa}

Some attacks are given the ability to control adversarial confidence defined in many ways, one of which is the $\kappa$ value, the logit gap between the adversarial class and the second most likely class.
Su \etal \cite{su2018robustness} empirically found that imposing a stricter
$\kappa$ constraint on CW attack slightly increases transferability on untargeted attacks. However, this measurement of adversarial confidence is scaling-sensitive because the $\kappa$ value scales if all parameters of the ReLU network is scaled, which is undesired because ReLU network is scaling-invariant. A straightforward extension of the $\kappa$ value is to compute the softmaxed logit gap between the adversarial class and the second most likely class, called \emph{Scaling-Insensitive $\kappa$ (SIK)} in this paper. By definition, scaling the logit will cause a much smaller change on SIK than SSK. The original $\kappa$ value is thus renamed \emph{Scaling-Sensitive $\kappa$ (SSK)}. For ease of understanding, SIK and SSK are mentioned collectively as $\kappa$ . To make AEs crafted from different algorithms and surrogates comparable, we measure SSK and SIK by direct calculation for all AEs. Then we divide them into two groups based on whether the AE
successfully fools the cloud.
Invoking the Bayesian rule:
$
  P(\text{success} \mid  \kappa)
  =P(\text{success}) \times P( \kappa \mid \text{success})/ P( \kappa)
  \propto P( \kappa \mid \text{success})/ P( \kappa)
$, we can see that the change in transfer success rate
is proportional to the change in $P( \kappa \mid \text{success}) / P(\kappa)$, a metric easier to compute.

\begin{figure}
  \centering
  \includegraphics[width=.8\linewidth]{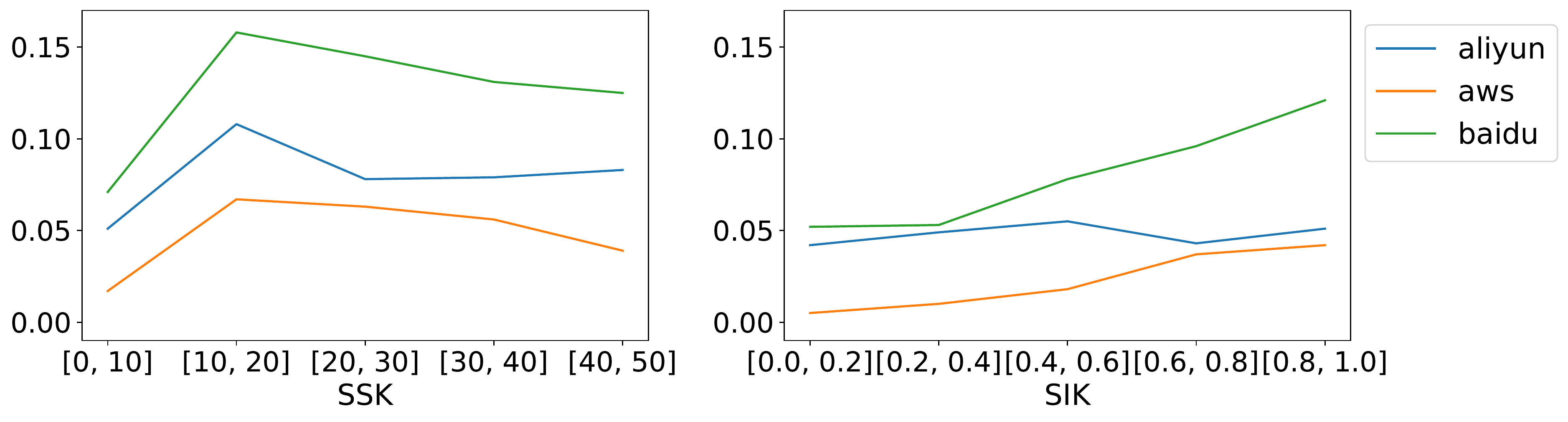}
  \caption{ $P(\text{success} \mid \kappa)$ across platforms when $\kappa$ is SSK and SIK, respectively. Both figures are scaled by a constant factor $P(\text{success})$. While all platforms show more vulnerability to a larger SIK, platforms except for Baidu do not show clear vulnerability to a large SSK.}
  \label{fig:kappa}
\end{figure}

Figure \ref{fig:kappa} plots $P( \kappa \mid \text{success}) / P(\kappa)$ for SSK and SIK. It indicates that the conclusion from Su \etal \cite{su2018robustness}
does not generalize as Aliyun and AWS do not show higher vulnerability to larger SSK. However, Figure \ref{fig:kappa} shows transferability
of AEs is nearly linear to SIK. This suggests that the conclusion from Su \etal \cite{su2018robustness}
is only a special case of a more general property. They might get their conclusion
because increasing SSK probably leads to a larger SIK. We further run a linear regression to quantify the impact, shown in Table \ref{tb:kappa_regression}.   The result shows $P(\text{SSK} \mid \text{success}) / P(\text{SSK})$ is not strongly explained by SSK ($p=0.303$), but $P(\text{SIK} \mid \text{success}) / P(\text{SIK})$ strongly depends on SIK ($p=0.002)$. Therefore, we conclude that SIK is a good choice for measuring adversarial confidence in transfer attack while SSK is not representative enough.

\begin{table}
  \centering
  \caption{ Linear regression results for SSK and SIK respectively. Adjusted R-squared are 0.653 for SSK regression and 0.821 for SIK regression.
    SSK and SIK are discretized as Figure \ref{fig:kappa} and revalued from one to five. For each entry, the upper is the result of SSK regression and the lower is of SIK regression.
  }
  \label{tb:kappa_regression}
  \resizebox{0.5\linewidth}{!}{
    \begin{tabular}{c >{}c >{}c >{}c}
      \toprule
      \textbf{IV}                  & \textbf{coefficient}    & \textbf{$p$-value}      & \textbf{5\% interval} \\ \midrule
      \multirow{2}{*}{is\_Alibaba} & 0.065                   & 0.003$\threeS$          & [0.026,0.103]         \\
                                   & 0.019                   & 0.050$\twoS$            & [0.000,0.039]         \\
      \multirow{2}{*}{is\_Baidu}   & 0.112                   & 0.000$\threeS$          & [0.073,0.150]         \\
                                   & 0.049                   & 0.000$\threeS$          & [0.030, 0.068]        \\
      \multirow{2}{*}{is\_AWS}     & 0.030                   & 0.117                   & [-0.009,0.068]        \\
                                   & -0.006                  & 0.482                   & [-0.026,0.013]        \\ \hdashline
      \multirow{2}{*}{SSK / SIK}   & \textbf{0.005}          & \textbf{0.303         } & [-0.005,0.015]        \\
                                   & \textbf{0.010         } & \textbf{0.002$\threeS$} & [0.004,0.014]         \\
      \bottomrule
      \multicolumn{4}{l}{\oneS  $p<.1$, \twoS  $p<.05$, \threeS  $p<.01$.}                                     \\
    \end{tabular}
  }

\end{table}

\begin{figure}
  \centering
  \begin{subfigure}{.32\linewidth}
    \centering
    \includegraphics[width=\linewidth]{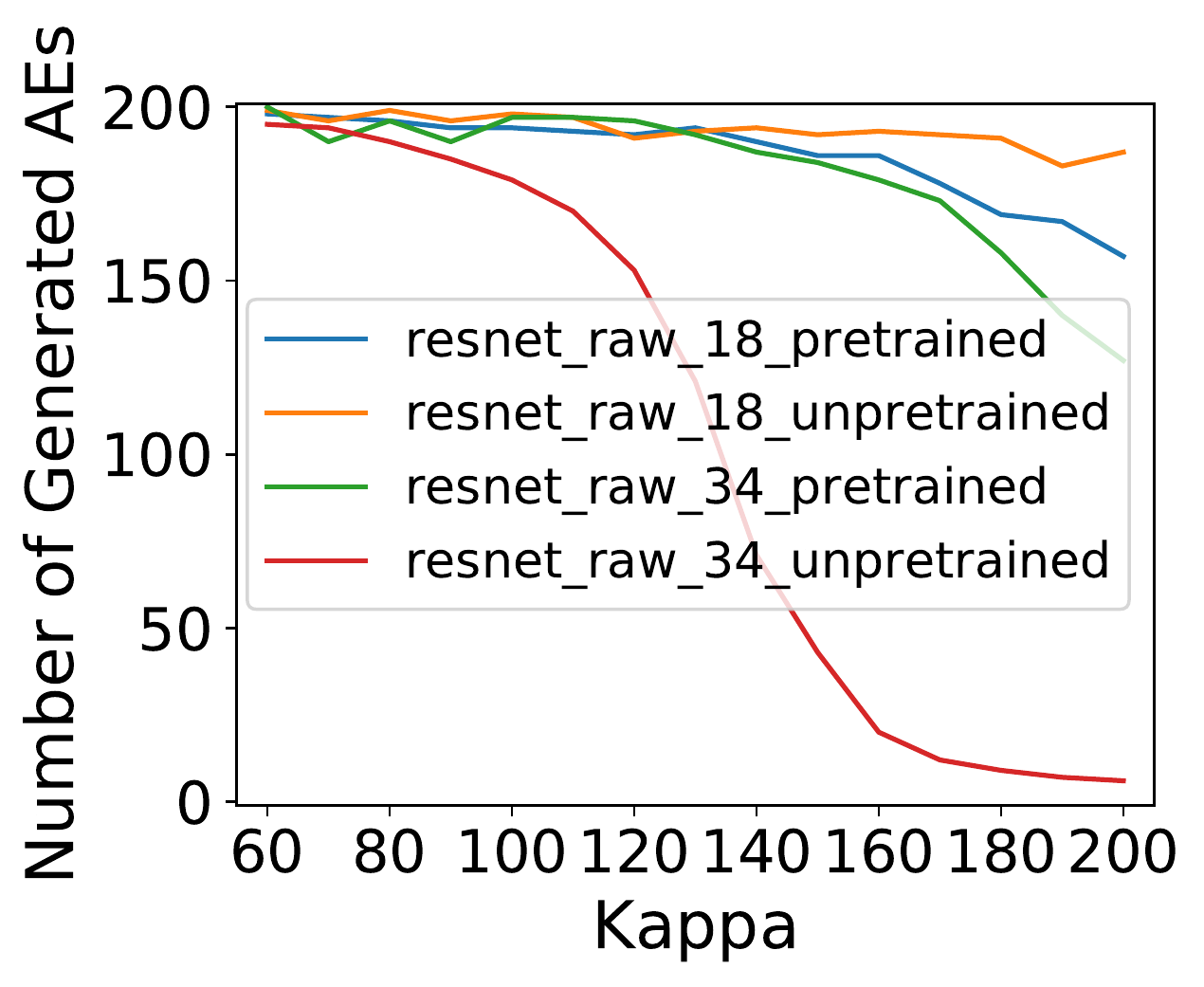}
    \caption{Number of AEs}
    \label{subfig:num_AE_kappa}
  \end{subfigure}
  \begin{subfigure}{.32\linewidth}
    \centering
    \includegraphics[width=\linewidth]{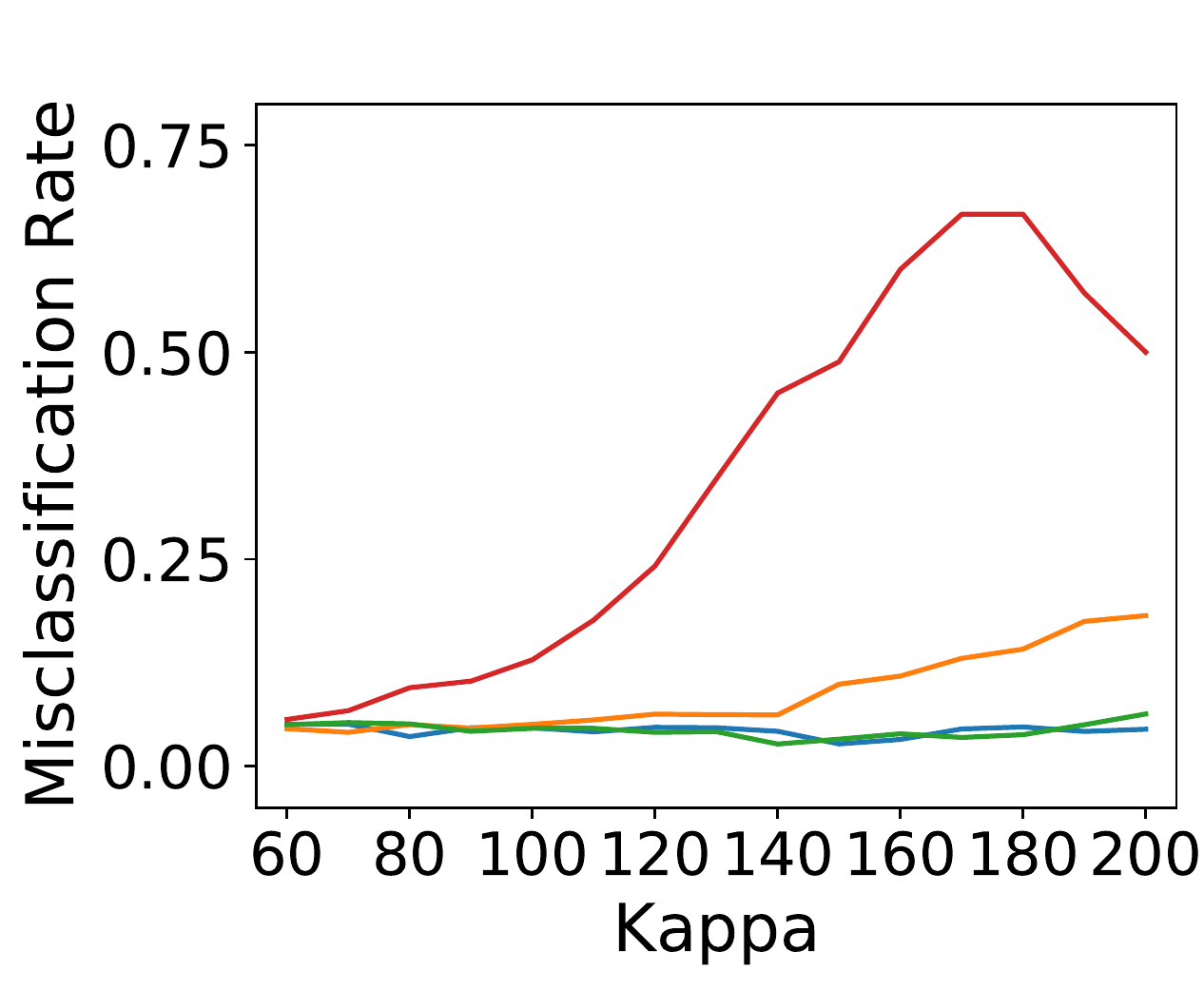}
    \caption{CW on AWS}
    \label{subfig:aws_CW_kappa}
  \end{subfigure}
  \begin{subfigure}{.32\linewidth}
    \centering
    \includegraphics[width=\linewidth]{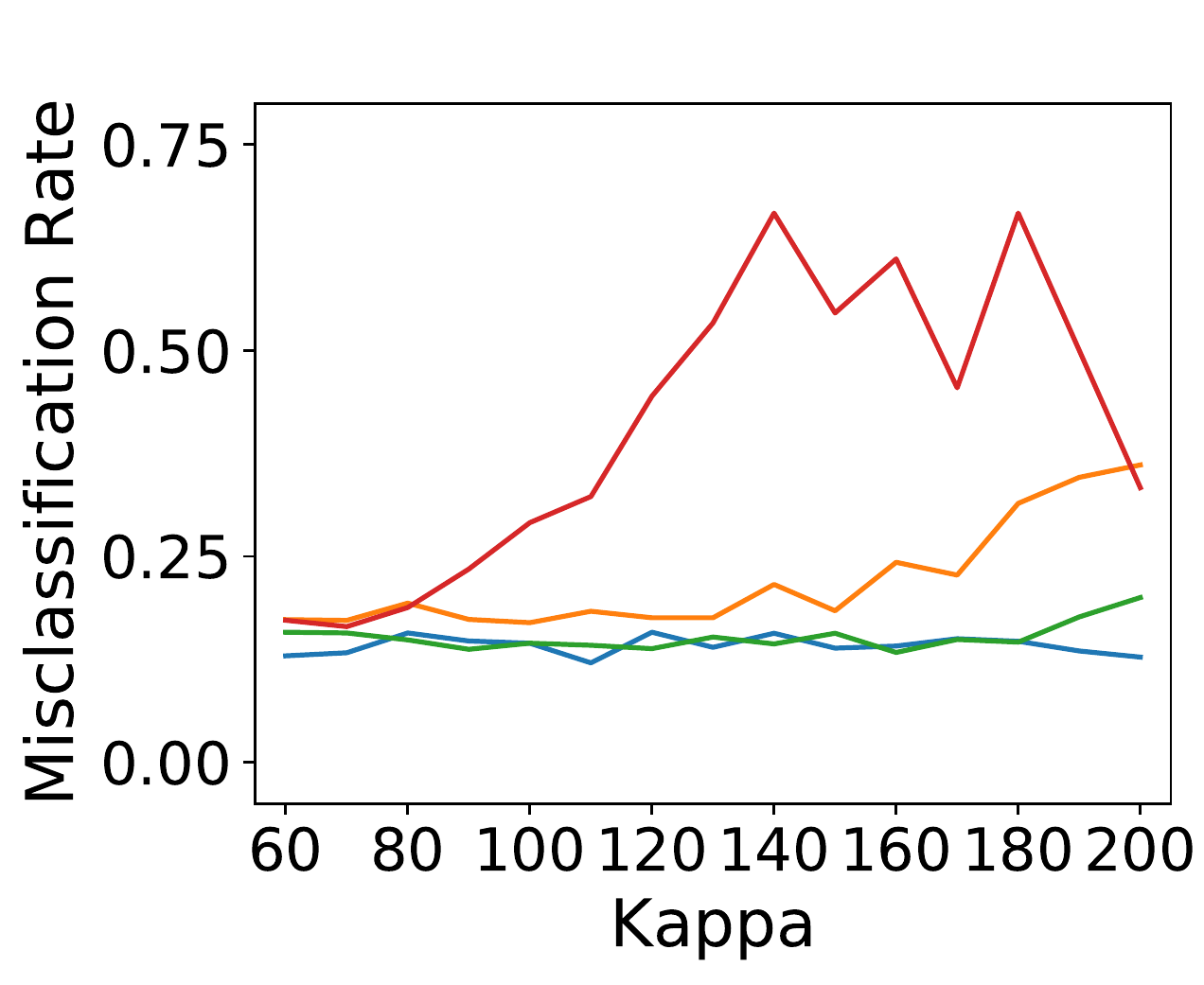}
    \caption{CW on Baidu}
    \label{subfig:baidu_CW_kappa}
  \end{subfigure}
  \caption{The misclassification rate for CW attack with different $\kappa$ thresholds, measured on AWS and Baidu using four different surrogates. The number of AEs drops because we use a fixed iteration budget, but for large kappa this budget no longer guarantees an adversarial perturbation.}
  \label{fig:CW_kappa}
\end{figure}

In practice, we want to know whether increasing the SSK for the CW attack is sufficient to increase the transferability as well. To answer this question, we perform CW attack with different $\kappa$ thresholds. The result is shown in Figure \ref{fig:CW_kappa}. We can see from Figure \ref{subfig:aws_CW_kappa} and Figure \ref{subfig:baidu_CW_kappa} that a larger SSK does not always increase transferability. For example, for the blue line in Figure \ref{fig:CW_kappa}, while the number of AEs drops at roughly 25\% when we increase SSK from 60 to 200, the transfer rate against the AWS platform does not improve.  Therefore, increasing SSK is not useful enough for a real transfer attack.

\begin{observation}
  The logit difference between the adversarial class and the second most likely class, called $\kappa$ value, is neither a good measurement for the transferability nor a good tool to increase the transferability.
\end{observation}

\subsubsection{Connection Between Transferability and Intrinsic Classification Hardness}
\label{sec:intrinsic_hardness}

\begin{figure}
  \centering
  \includegraphics[width=.8\linewidth]{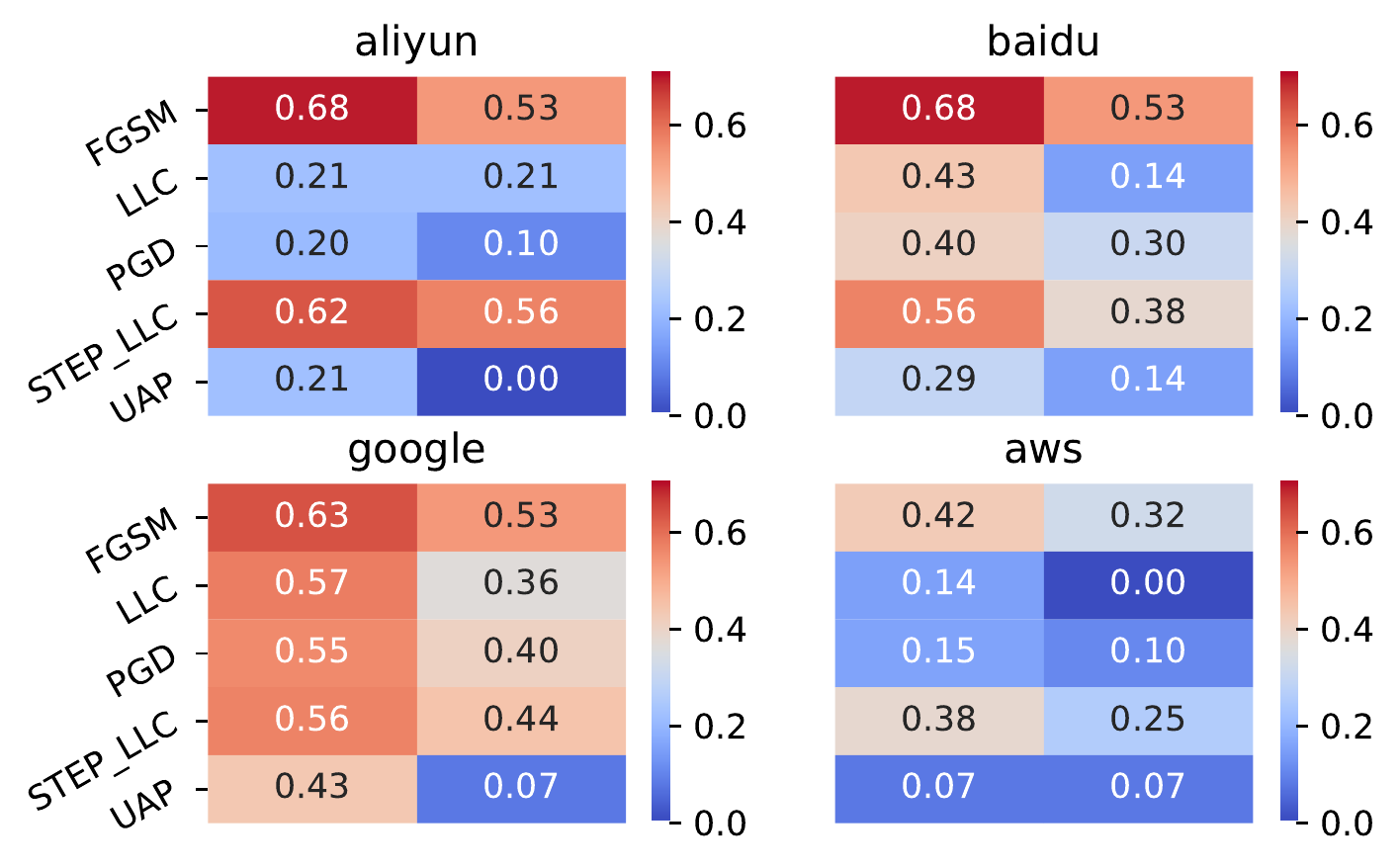}
  \caption{Comparison of the transferability between AEs generated from natural AEs and normal seed images. For each plot, the left column is the misclassification rate of AEs generated from natural AEs and the right column is the misclassification rate of AEs generated from other seed images. It shows that the left column is always greater or equal to the right column.}
  \label{fig:natural_AE_vs_normal}
\end{figure}

An image is defined to be a \emph{natural} AE w.r.t. a surrogate model if it can fool the surrogate without any disguise. When conducting transfer attack in the
real world, natural AEs are unavoidable since the surrogate cannot achieve 100\% accuracy. We are thus motivated to find out if perturbing natural AEs can provide better transferability.

We compare the transferability of AEs generated from natural AEs with AEs generated from other seed images in Figure \ref{fig:natural_AE_vs_normal}. The natural AEs disguised by adversarial perturbation achieve higher success rate than AEs generated from other seed images consistently. It suggests that the intrinsic classification hardness is important in deciding transferability and real world attackers should prefer natural AEs in transfer attacks.
The result is intuitive because natural-AE seeds are believed to be closer to the theoretical classification border and an adversarial perturbation is more likely to transfer on these images.

\begin{observation}
  AEs generated from seed images that are hard to classify on surrogates have better transferability.
\end{observation}

\section{Discussion}
\subsection{Ethic}
When evaluating attacks on the platforms, we conduct normal queries with the official API and pay the usage fee as required. Therefore, our evaluations are legitimate and do no harm to the  operations of the cloud services. The related companies are informed about our experiment to exclude any potential harm, \eg, do not use the uploaded images in any form. We have received conﬁrmation from AWS and Baidu.

\subsection{Limitation and Future Work}

 Our goal is to provide a systematic view of factors that affect transferability. Therefore, we leave some conclusions not fully understood. For example, we do not validate the hypothesis that the number of the local optimal increases exponentially when the task complexity increases. These arguments are beyond the scope of this paper and require further study in the future. In addition, we do not find the best hyperparameter setting in the real settings. Although these answers are useful, it requires massive resources which we cannot afford, thus left for future work. In particular, our conclusions about the attack algorithms are specific to the chosen hyperparameters, not representing their ability under different hyperparameters. For example, we set $\kappa=0$ for CW attack, but choosing a large $\kappa$ may be preferable for attackers. In our experiment, however, we set $\kappa=0$ for a fair comparison of CW with other attacks which cannot manipulate kappa.

\section{Related Works}
Early works on adversarial attacks \cite{szegedy2013intriguing, goodfellow2014explaining, carlini2016evaluating,  moosavidezfooli2015deepfool, aleks2017deep, tramr2017ensemble} mainly focus on white-box attacks, where the adversary has full access to the target model.
Considering the lack of information in black-box attacks, efforts have been made in two directions: 1) \textit{query-based attacks} that estimate the gradient information used to facilitate the generation of AEs\cite{chen_zoo_2017, ilyas_black-box_2018} and 2) \textit{transfer attacks} that generate AEs using a surrogate model and expect them to transfer to the unknown target model \cite{papernot_transferability_2016, papernot_practical_2017}. Our paper focuses on transfer attacks.



The transferability of AE in deep learning models was first discovered by Szegedy \etal \cite{szegedy2013intriguing}. Papernot \etal \cite{papernot_practical_2017, papernot_transferability_2016} utilized this property to perform black-box attack to real MLaaS systems, which proves its possibility for real attacks. Carlini \etal \cite{carlini2016evaluating} proposed that high-confidence adversarial examples increase the transferability.

Efforts have been made to understand transferability of AEs.
Liu \etal \cite{liu2016delving} found that unlike untargeted AEs, targeted AEs almost never transfer and proposed to use an ensemble of surrogates to increase transferability of targeted attacks. Wu \etal \cite{wu_understanding_2018} studied influencing factors of transferability and point out that reduced variance of gradients results in better transferability. Su \etal \cite{su2018robustness} conducted a comprehensive analysis of transfer attack using 18 different surrogates. They found that relaxing norm constraint of adversarial perturbation generates better transferability. They concluded that FGSM $>$ PGD $>$ CW in the sense of transferability as well. Demontis \etal \cite{demontis2018adversarial} further highlighted that simpler surrogates and better gradient alignment provide better transferability. This conclusion was based on adding different levels of regularization and found stronger regulation provides better transferability. In addition, they concluded that targets which have large gradients for the input are more vulnerable to transfer attack as well.  Our work aims at examining whether their conclusions generalize to transfer attack in the real applications and extending their results to get more insights. In particular, when evaluating the relation between surrogate complexity and transferability, we believe that a better way to examine the impact of complexity in the real settings is to directly use surrogates with different depths but the same architecture family, rather than changing regularization. Our conclusion about the non-monotonicity of the effect of surrogate depth on transferability is a complement to Demontis \etal.

\section{Conclusion}

In this paper, we identify the difficulties of evaluating transferability of adversarial example (AE) in the real world and propose customized metrics to address these difficulties. Based on the proposed metrics, we conduct a systematic evaluation of real-world transfer
attacks on four popular MLaaS systems, Aliyun, Baidu Cloud, Google Cloud Vision and AWS Rekognition. The evaluation leads to the following new conclusions on top of the existing ones made in lab settings:
\textbf{(1)} model similarity concept is ill-suited for transfer attack;
\textbf{(2)} surrogates of suitable complexity can exceed simpler and deeper counterparts;
\textbf{(3)} MLaaS systems have different level of robustness to transfer attack and could use more effort to improve;
\textbf{(4)} Strong adversarial algorithms do not necessarily transfer better, and single-step algorithms transfer better than iterative algorithms;
\textbf{(5)} adversarial algorithm and the target platform are the most important factors for transfer attacks, and the most beneficial practice is to choose an appropriate adversarial algorithm;
\textbf{(6)} no dominant surrogate architecture exists in the real transfer attack.
\textbf{(7)} large $L_2$ norm of adversarial perturbation can be a more direct source of transferability than $L_\infty$ norm.
\textbf{(8)} larger gap between the posterior of logits makes better transferability.
\textbf{(9)} classification hardness is  preferred when choosing seed images of transfer attack. 

\section{Acknowledgement}

We would like to thank our shepherd and the anonymous reviewers for their valuable suggestions.
This work is partly supported by the Zhejiang Provincial Natural Science Foundation for Distinguished Young Scholars under No. LR19F020003, NSFC under No. 62102360, and U1836202, and the Fundamental Research Funds for the Central Universities (Zhejiang University NGICS Platform). Ting Wang is partially supported by the National Science Foundation under Grant No. 1951729, 1953813, and 1953893.

\bibliographystyle{acm}
\bibliography{reference}

\section{Appendix}

\subsection{Validity and Bias of Equivalence Dictionaries}
\label{appendix:dict_justify}
An equivalence dictionary consists of different label mappings.
Each label mapping can be viewed as a new class for a MLaaS model without making critical changes.

However, building a good equivalence dictionary is important for getting the real
performance. Although we use human knowledge to create it, bias can be introduced.
We formally discuss the level of bias introduced using the following
proposition.
\begin{proposition}
  \label{prop:bias}
  Suppose $\mathcal{D}$ contains all acceptable MLaaS classes for a local class $c$. $M$ is the class mapping we construct.
  If $M$ satisfies $M \subset \mathcal{D}$, then the measured matching rate is lower and the measured misclassification rate is higher than their true value. If $\mathcal{D} \subset M$, then the measured matching rate is higher and
    the measured misclassification rate is lower than their true value.
          When $\|\mathcal{D}-M\|$ shrinks to empty set, the bias decreases to zero as well.
\end{proposition}

The proof of Proposition~\ref{prop:bias} is straightforward based on its definition.
To make $\|\mathcal{D} - M\|$ as small as possible, we construct the equivalence dictionaries as described in Section~\ref{sec:metric}.
This method is good at capturing acceptable predictions in practice,
reducing the number of missed acceptable predictions to a very low level, \ie,
$\|\mathcal{D}-M\|$ is nearly empty.
In our experiment, we find some class mappings include super-classes, such as including \emph{sports} in the class mapping of \emph{baseball},
which makes $\mathcal{D} \subset M$.
As shown in Proposition \ref{prop:bias}, this leads to a larger matching rate and smaller misclassification rate, which is potentially destructive to our evaluations.
However, as we see from the results, all the matching rates are small and most of the misclassification rates are significant, indicating that the bias should be of little harm to the conclusions we make.

\subsection{Impact of Irrelevant Factors on the Conclusions} \label{sec:regression}

Traversing all the possible factor settings in transfer attacks is expensive. In fact, to ensure balanced observations, the experiment settings are $\Omega = \Omega_1\times \Omega_2 \times \dots \times \Omega_n$, where $\Omega_i$ is the setting space of one factor, \eg, pretraining, surrogate dataset, adversarial algorithm, \etc, and $n$ is the total number of factors. This means the total setting space is exponential.
The large setting space makes it extremely hard to derive conclusions that are generalizable in a reasonable sense.

We address this problem by taking empirical expectation over the irrelevant setting space, thus minimizing the influence of the specific settings of irrelevant factors. To be exact, we conclude the influence of one specific factor by averaging over settings with this  factor fixed and other factors varied in a balanced way. This is realized by running regression on balanced data which can be viewed as grid samples from $\Omega$. For example, 
when concluding the effect of a target platform, the setting space of the target platform is the considered factor space (denoted as $\Omega_{i^*}$), and the setting space of pretraining, dataset and surrogate model, \etc, are the irrelevant space $\prod_{i\ne i^*} \Omega_i$.
The regression methodology  provides automatic empirical expectation over the irrelevant settings based on the provided data, and we use balanced data in regression to further make the expectation taken on $\prod_{i\ne i^*} \Omega_i$ to be unbiased.
Therefore, the effect of a specific factor is in fact eliminated by taking expectation. In other words, the conclusions made on one factor, \eg, pretraining, are not affected by the specific settings of other factors, \eg, adversarial algorithm. Since what we actually do is an empirical expectation, the effect of factors might not be totally eliminated, but still minimized.

\subsection{Correctness and Generalization of Applying OLS Analysis}
\label{appendix:ols_correlation}

In multi-factor analysis, OLS analysis is commonly adopted to decompose the influence of each factor. However, due to its simplicity, it sometimes only reflects some aspects of the underlying influence function.

First, since the underlying influence of factors might be non-linear, the result of OLS which is linear is not suitable to be taken as the ground truth of the underlying influence function. The correct application is to use the ``direction'' and ``size'' of impact, \ie, whether the impact is positive and how large the impact is, which is the central point of our observations and analyses. However, the reader should note that these directions are local and are technically restricted to the applied linear model.

Second, the OLS analysis finds ``relationship'' from the data and is a correlation analysis naturally. This means that all results are ``correlations'' but not ``causation'' unless there are additional experiments targeting the causation. In our paper, the only causation analysis is in Section \ref{sec:correlation} where we generate random AEs with largest $L_2$ and fixed $L_\infty$ to test the influence of $L_2$ norm.

Third, the $p$-values are computed based on the assumption that the residual of linear model is Gaussian-distributed. In practice, this assumption may not hold, leading to biased $p$-values. Specifically, in our experiment, we find that the residual is roughly $t$-distributed with a degree of freedom 16 which is non-Gaussian. However, fixing this issue using techniques such as Cox-Box transformation\cite{Cox-Box} is ill-suited for binary variables and makes the result difficult to follow. In addition, for large degrees of freedom, t-distribution converges to Gaussian distribution. Therefore, since it has little bias on the result, we use the original OLS for clarity.

\subsection{Detailed Explanations for OLS Tables}
\label{appendix:ols_explanation}

Due to the page limit of the main text, we do not include detailed explanations on how to read the formatted OLS tables in Section \ref{sec:threat_setting_factor}. This section provides these materials for readers unfamiliar with OLS analysis.


To conclude a factor's impact on the dependent variable (called response variable as well), one need to look at all regressions on the dependent variable that include this factor.  If the coefficient changes dramatically when an extra factor is included, it usually suggests that strong interaction exists between this factor and the extra factor. If the coefficient is not changed when other factors are included, it means that the interaction is weak, if exists. When strong interaction is detected, it is preferable to perform another regression and include the interaction term, \eg, \emph{is\_pre$\times$is\_adv}, simplified as \emph{is\_pre$\times$adv} in this paper. This kind of technique, known as hierarchical regression, provides more information about the impact of factors.

The tables are formatted to include and directly compare results of multiple OLS regressions. Since Table \ref{tb:imagenet_regression} and Table \ref{tb:gender_regression} are formatted in the same way, we use Table \ref{tb:imagenet_regression} as an example. In Table \ref{tb:imagenet_regression}, regression $\mathbb{A}$ to $\mathbb{E}$ regress misclassification rate and regression $\mathbb{F}$ to $\mathbb{J}$ regress matching rate on the corresponding variables. Each regression from $\mathbb{A}$ to $\mathbb{E}$ includes different regressors. For one regression, if coefficients and standard deviations are provided in the table for a regressor, then it means that this regressor is included in this regression and the corresponding result is the provided value. Otherwise, if the space is left blank, it means that this regressor is not included in the regression. For example, in regression $\mathbb{A}$, the regressors are platform factors (from \emph{is\_Google} to \emph{is\_Aliyun}). In regression $\mathbb{B}$, the regressors are platform factors, pretraining factors (\emph{is\_pretrained}) and dataset factors (\emph{is\_adversarial} and \emph{is\_augmented}).

To avoid randomness introduced by the data on the result, $p$-values are provided. If a coefficient has a small $p$-value, highlighted by asterisks, it means that this coefficient is significantly different from zero, \ie, the impact is ``real''. In practice, only values highlighted with asterisks are considered as important and worth attention. Apart from $p$-values, coefficients of the OLS regression and their standard deviations are provided in the table. A positive coefficient indicates that this factor has a positive impact on the dependent variable and a negative coefficient indicates the opposite. In addition, a larger absolute value of the coefficient indicates a larger effect. The standard deviation represents how much uncertainty is involved in computing the coefficient, and is mainly used as a supplement to $p$-values. Details about the OLS regressions can be found in the released code repository.

\subsection{Further Discussion about $L_\infty$ and Misclassification Rate}
\label{sec:infty-mis}

From Figure \ref{fig:correlation} in Section \ref{sec:correlation}, we have shown that $L_2$ has large correlation with misclassification rate while the correlation between $L_\infty$ and misclassification rate is relatively small. Since $L_2$ and $L_\infty$ has a correlation 0.51, it is possible that even this roughly 0.4 correlation between $L_\infty$ and misclassification rate is largely due to the dependence between $L_2$ and $L_\infty$. We prove this claim in this discussion.

For notational reasons, we define $X_1$ to be $L_2$ norm  and $X_2$ to be $L_\infty$ norm,  divided by their standard deviation, respectively.  We further define $Y$ to be the misclassification rate divided by its standard deviation. Thus, $\var{X_1} = \var{X_2} =\var{Y}=1$ and the correlation between them are equivalent to the covariance. Let $\epsilon = X_2 - X_1\times \cov(X_1, X_2)$. Since $\cov(\epsilon, X_1) = \cov(X_1, X_2) - \cov(X_1\times \cov(X_1, X_2), X_1) = 0$, we know $\epsilon$ is actually uncorrelated with $X_1$. Therefore, $ X_2 = \epsilon + X_1\times \cov(X_1, X_2)$ decomposes the effect of $X_1$ and $\epsilon$ is the ``$L_2$-free'' term of $L_\infty$. Now we claim that this $L_2$-free term has a very small correlation to misclassification rate. Indeed, $\cov(\epsilon, Y) = \cov(X_2, Y) - \cov(X_1\times \cov(X_1, X_2), Y) = \cov(X_2, Y) - \cov(X_1, X_2) \times \cov(X_1, Y)$. Plug in all the correlation values in Figure $\ref{fig:correlation}$, we get the correlation between $\epsilon$ and $Y$ is $-0.069$ for Aliyun, $-0.0439$ for Baidu, $-0.0633$ for Google and $-0.0682$ for AWS. Therefore, the $L_2$-free contribution of $L_\infty$ to the misclassification rate is slightly negative, almost negligible.

\subsection{Guidelines for Defending Transfer Attacks}
\label{sec:guideline}
\begin{itemize}
  \item (Observation \ref{obs:algorithm}) Use FGSM first when evaluating the robustness to transfer attacks. 
  \item (Section \ref{sec:norm} and \ref{sec:kappa}) As long as the attack algorithm is able to control adversarial confidence (SIK) and perturbation norm of AEs, use a large SIK value and a large $L_2$
        perturbation norm for better transferability.
  \item (Section \ref{sec:intrinsic_hardness}) Maintain a pool of images that are intrinsically hard to classify.
        Use these images as seeds to test the robustness of the target model.
\end{itemize}

\subsection{Parameter Settings for White-box Attacks and Augmentation}
\label{appendix:params}
The detailed settings of white-box adversarial attacks on the surrogate models are shown in Table \ref{tab:params}. We set kappa to zero for CW2 to make it comparable to other attacks, and it achieves a 100\% success rate on surrogates for all cases. The iteration times of iterative attacks, though might improve the attack when increased further, are sufficient in our experiments because all of them reach the $\epsilon$ bound, as shown in Figure \ref{fig:LINF}. The choice of specific $\epsilon$ bound is based on the principle that an attacker would like to set $\epsilon$ as large as possible while maintaining visual quality, \ie, being unnoticed by human, because it would lead to better transferability. Since visual quality is hard to quantify, we try different values for $\epsilon$, and decide that $0.1$ is a good choice for ImageNet under the condition that human eyes should not perceive the difference. Actually, with $\epsilon=0.1$, there are already some mild noise patterns that are noticeable to human eyes. Therefore, our experiments can be viewed as a stress test of MlaaS systems.

\begin{table}[htbp]
  \centering
  \caption{Hyperparameter setting of adversarial algorithms. $\epsilon$ is all set to be 0.1 for ImageNet and Adience.}
  \resizebox{0.5\linewidth}{!}{
    \begin{tabular}{p{5em}p{8.5em}}
    \hline
    \textbf{Attacks} & \textbf{Configuration} \\
    \hline
    BL-BFGS & initial const=0.01 \newline search steps=10\newline max iterations=40 \\
  
    CW2   & $\kappa$=0\newline{}initial const=0.001\newline{}learning rate=0.02\newline{}max iterations=100 \\
    
    DeepFool & max iterations=100\newline{}overshoot=0.02 \\
    
    
    PGD   &  number of steps=10 \\
    
    RFGSM &  $\alpha$=0.5 \\
    
    
    \hline
    \end{tabular}%
  }
 \label{tab:params}%
\end{table}%

The detailed setting of data augmentation is described below. Color jitter has brightness=0.1, contrast=0.1 and saturation=0.1, which means the augmented image will be in the range of these deviations. Random affine has maximum degree 30 and a maximum translate (0.1, 0.1). Random rotation has a maximum degree 30. Random horizontal flip, random vertical flip and random perspective have no parameter. They are applied with a probability of 0.5 independently. After finishing an epoch of training, the augmentation is redone, which means the dataset is enlarged by the number of epoch times.
In our case, it is 50.

\subsection{Implementation Changes on DEEPSEC}
\label{appendix:deepsec}
The original DEEPSEC implementation presented by Ling \etal \cite{ling_deepsec_2019} has issues about numerical stability which have been fixed by the original authors. We apply this new version of DEEPSEC in our experiments. Furthermore, we removed the random initialization in the PGD \href{https://github.com/kleincup/DEEPSEC/blob/master/Attacks/AttackMethods/PGD.py}{code}, which is proposed in RFGSM, to make it directly comparable to FGSM. This action is not a fix to DEEPSEC but specific to our usage, and users are still encouraged to use DEEPSEC but not our changed version. Apart from some auxiliary code, what we include in our repo is mostly the test code which sends the data to MLaaS and analyzes the feedback.

\subsection{Discussion about the Cutting Threshold}
\label{appendix:threshold}

As discussed in Section \ref{sec:metric}, we address the multiple predictions problem by applying a cutting threshold estimated on the returns of original images. In fact, it is worthwhile to consider whether these estimated thresholds are applicable to other datasets and whether real users will apply the threshold cutting mechanism. We discuss these two issues below, respectively.

First, the estimation is conditioned on the chosen classes and no refinement can be made because no knowledge about the ``real'' threshold is known. However, we estimate the thresholds on the used dataset and apply it on this dataset. Therefore, the threshold does not need to ``transfer''. In addition, as stated in Section \ref{sec:metric}, the choice of thresholds only affects the comparison of MLaaS systems which is limited to the discussed tasks but leaves the discussion of other factors unharmed. Therefore, although the estimation might be biased, it does not affect conclusions about other factors but only platform factors.
For attacks with seed images from undiscussed classes, they should do the threshold searching again and not directly apply our thresholds.

Second, the threshold cutting mechanism is a reasonable way to be adopted by users of MLaaS systems since the number of highly confident labels may vary.
Furthermore, if the users apply threshold cutting, then the chances are that the thresholds derived from our methods are good estimations for the applied thresholds. This is because these users would like to focus on a few predictions without significant accuracy drop, which is aligned to our principles. Therefore, our conclusions provide an important insight of transfer attack for these users.

\subsection{Discussion about Input Gradient Size} \label{appendix:input_grad}
We would like to bring back that the input gradient size on our target models, MLaaS platform models, is not available. Nevertheless, in order to explore the relation between transferability of AEs and their input gradient size on the target model, we build up some local models to simulate the process of transfer attacks, i.e., use one local model as the target model and the others as the surrogate models. The experiments are conducted on the Adience dataset, and we choose the classic pretrained VGG-16 model as the target model. Then we use AEs generated on other local models to attack the target model and measure their input gradient size on the target model. Note that these AEs are generated with multiple model architectures, multiple surrogate dataset settings and multiple adversarial attack methods, following the evaluation settings in Section \ref{sec:kappa}. Specifically, the surrogate datasets include raw and augmented datasets; the surrogate models may be trained with or without the adversarial training; the surrogate models' architectures include ResNet, VGG-16 and Inception V3; depths of the ResNet include 18, 34 and 50; the adversarial attack methods include BL-BFGS, CW2, DeepFool, FGSM, LLC, PGD, RFGSM, STEP-LLC and UAP.

\begin{figure}
	\centering
	\includegraphics[width=0.6\linewidth]{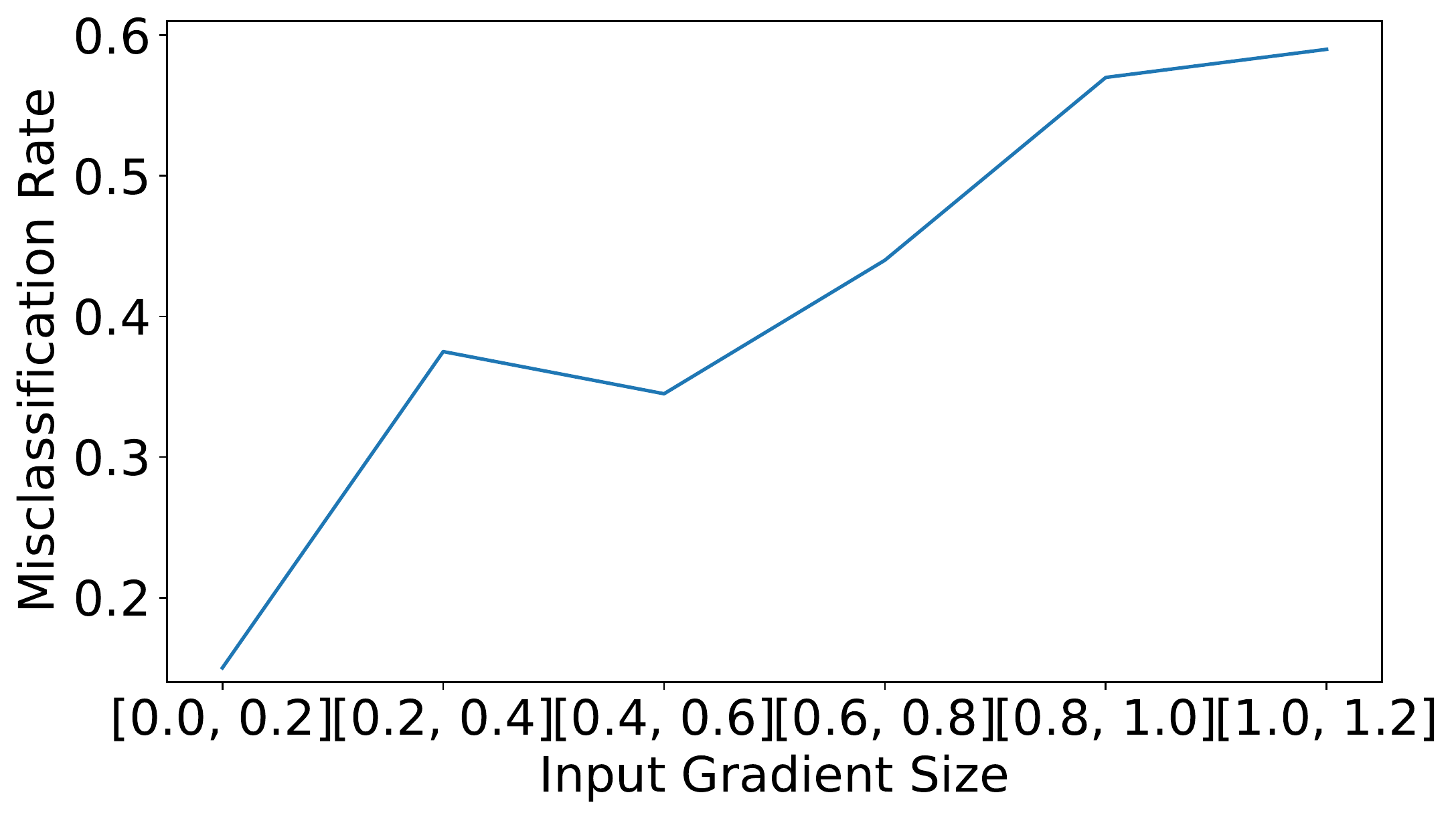}
	\caption{The relation between transferability and the input gradient size on the target model.}
	\label{fig:input_grad_size}
\end{figure}

The experiment results are shown in Figure~\ref{fig:input_grad_size}. It is evident that AEs with larger input gradient size on the target model tend to have better transferability, regardless of how the AE is generated, which is consistent with the conclusion of Demontis \etal \cite{demontis2018adversarial}.

\end{document}